\documentclass[runningheads]{llncs}
\usepackage[T1]{fontenc}
\usepackage{graphicx}
\usepackage{booktabs}
\usepackage[misc]{ifsym}

\usepackage{mwe}
\usepackage{amsmath}
\usepackage{amssymb}
\usepackage{tabularx} 
\usepackage{array}  
\usepackage{multirow}
\usepackage{graphicx}
\usepackage{caption}
\usepackage{subcaption}
\usepackage{float}
\usepackage{makecell}
\usepackage{hyperref}

\usepackage{amsmath,amsfonts,bm}

\def\vzero{{\bm{0}}}

\def\vtheta{{\bm{\theta}}}

\def\mI{{\bm{I}}}

\def\rvepsilon{\boldsymbol{\epsilon}}

\def\rvmu{{\bm{\mu}}}

\def\rvs{{\mathbf{s}}}

\def\rvv{{\mathbf{v}}}

\def\rvx{{\mathbf{x}}}

\def\rvz{{\mathbf{z}}}

\begin{document}

\title{Loss Functions in Diffusion Models: A Comparative Study}
\tocauthor{Dibyanshu Kumar, Philipp Väth, Magda Gregorová}
\toctitle{Loss Functions in Diffusion Models: A Comparative Study}

\titlerunning{Loss Functions in Diffusion Models: A Comparative Study}

\author{Dibyanshu Kumar (\Letter)\orcidID{0009-0007-2542-4781} \and \\
Philipp Väth\orcidID{0000-0002-8247-7907} \and \\
Magda Gregorová\orcidID{0000-0002-1285-8130}}

\authorrunning{D. Kumar et al.}

\institute{Center for Artificial Intelligence and Robotics, Technical University of Applied Sciences Würzburg-Schweinfurt, Franz-Horn-Straße 2, Würzburg, Germany\\
\email{kumardibyanshu05@gmail.com, philipp.vaeth@thws.de, magda.gregorova@thws.de}}

\maketitle              

\begin{abstract}
Diffusion models have emerged as powerful generative models, inspiring extensive research into their underlying mechanisms.
One of the key questions in this area is the loss functions these models shall train with. 
Multiple formulations have been introduced in the literature over the past several years \cite{ho2020denoising,song2019generative,kingma2021variational,salimans2022progressive} with some links and some critical differences stemming from various initial considerations.
In this paper, we explore the different target objectives and corresponding loss functions in detail. We present a systematic overview of their relationships, unifying them under the framework of the variational lower bound objective.
We complement this theoretical analysis with an empirical study providing insights into the conditions under which these objectives diverge in performance and the underlying factors contributing to such deviations.
Additionally, we evaluate how the choice of objective impacts the model’s ability to achieve specific goals, such as generating high-quality samples or accurately estimating likelihoods.
This study offers a unified understanding of loss functions in diffusion models, contributing to more efficient and goal-oriented model designs in future research.

\end{abstract}

\section{Introduction}
\label{sec:intro}

Diffusion models \cite{ho2020denoising} have become a cornerstone of generative modeling in recent years, demonstrating remarkable capabilities in generating high-quality data. 
Given a sample $\rvx_0 \sim q(\rvx)$ from a data distribution, the forward process in diffusion models incrementally corrupts the data by adding small amounts of Gaussian noise over multiple steps $T$. 
This process is defined as $q(\rvx_t \mid \rvx_{t-1}) = \mathcal{N}(\rvx_t; \alpha \rvx_{t-1}, \sigma^2 \mI)$, where \( \alpha \) controls the scaling of the data $\rvx_{t-1}$, and \( \sigma \) controls the magnitude of the added noise. 
The objective is then to learn the reverse process $q(\rvx_{t-1} \mid \rvx_t)$ which enables the generation of new samples by starting from pure Gaussian noise $\rvx_T \sim \mathcal{N}(\vzero, \mI)$ and iteratively denoising it to recover realistic data. 
This framework of probabilistic modeling allows diffusion models to capture complex data distributions, making them highly effective for a wide range of generative tasks.

The class of diffusion models has seen several notable contributions, particularly in the development of training objectives. 
In score-based modeling \cite{song2019generative}, the reverse process is learned by minimizing a denoising score-matching objective. 
Ho et al. \cite{ho2020denoising}, in their work on DDPM, generated high quality images by adopting noise prediction $\rvepsilon$ as the primary objective. Variational Diffusion Models (VDM) \cite{kingma2021variational} used the training objective, formulated in terms of the Signal-to-Noise Ratio (SNR), which achieved the best likelihood estimation. 
Additionally, the authors of Progressive Distillation \cite{salimans2022progressive} modeled the rate of change in data distribution over time, presenting a novel loss function that combines the data representation $\rvx$ and the noise component $\rvepsilon$. 
This objective was instrumental in reducing the number of sampling steps required to generate high-quality samples. These advancements highlight the critical role of loss function design in improving the performance and efficiency of diffusion models.

Existing research has explored the theoretical equivalence of various training objectives used in diffusion models. For instance, \cite{song2020score} established connections between score matching and diffusion-based generative frameworks by leveraging stochastic differential equations to model the forward process, thereby aligning it with continuous distributions that evolve over time. 
Similarly, \cite{kingma2021variational} introduced the Evidence Lower Bound (ELBO) objective for diffusion, inspired by Variational Autoencoders \cite{kingma2019introduction}. 
More recently, \cite{kingma2024understanding} demonstrated that diffusion model objectives are fundamentally equivalent and closely related to the ELBO framework. 
However, while these works highlight the theoretical equivalence of the loss functions, they lack a structured analysis of their formulations under a single framework. 
Furthermore, there is no empirical study investigating whether the mathematical equivalence between objectives persists when training diffusion models with deep neural networks. 
Therefore, there is only limited understanding of how these loss formulations differ in terms of performance. This gap highlights the need for a systematic exploration of outcomes of these theoretical connections.

In this study, we conduct a comprehensive comparison of different training objectives, specifically the weighted and the ELBO objectives, formulated for four different target predictions of the diffusion models: data $\rvx$ , noise $\rvepsilon$, rate of change in the data distribution $\rvv$ and score $\rvs$. 
We derive the negative ELBO loss in terms of these targets and establish mathematical relationships with the most commonly used diffusion loss functions. 
These relationships help us to design experiments that evaluate whether the theoretical equivalence between these objectives holds in practice when used for training over the same datasets. 
Our experiments highlight the differences and similarities in the theoretical foundations and practical behavior of these loss functions, particularly in terms of loss convergence during training and the quality of generated samples.
We explore the loss behavior across different diffusion timesteps, providing insights into the mechanisms that drive their performance and functionality. 
Additionally, we compare the outcomes of these training objectives in terms of data density estimation and sample quality, offering a comprehensive understanding of their roles in optimizing diffusion models. 

The paper is structured as follows: section \ref{sec:Model} provides the background on diffusion models. 
In section \ref{Loss_forms} we introduce the various target predictions used in diffusion models, derive the loss functions under different framework, and show the relation between them.
In section \ref{sec:Experiments} we describe the experiments we perform and give insights on the results obtained. Finally, we conclude by summarizing our findings and suggesting directions for future research. The code used in this study is available at: \url{https://github.com/dibyanshu100/LFDM}.

\section{Model}
\label{sec:Model}

In this section, we provide an overview of the forward and reverse processes used in diffusion models.

\subsection{Forward diffusion process}
\label{subsec:ForwardDiff}

The forward process in diffusion models is a Markov process, where the information in a given data $\rvx$ is progressively destroyed by adding noise in a series of timesteps, producing intermediate latent variables, denoted as $\rvz_t$, where $t \in [0,1]$ represents the corresponding timestep.
To achieve this, a schedule is used to define the amount of noise to be added and the signal to be removed at each timestep, regulated by parameters $\alpha_t$ and $\sigma_t$.
The distribution of latent variables and the Markov transition distribution in the forward process is defined as follows:

\begin{equation}
\begin{aligned}
q(\rvz_t \mid \rvx) &= \mathcal{N}(\rvz_t; \alpha_t \rvx, \sigma_t^2 \mI) \\
q(\rvz_t \mid \rvz_s) &= \mathcal{N}(\rvz_t; \alpha_{t|s} \rvz_s, \sigma_{t|s}^2 \mI)
\end{aligned}
\end{equation}

where $0 \leq s \leq t \leq 1$, $\alpha_{t \mid s} = \frac{\alpha_t}{\alpha_s}$ and $\sigma_{t \mid s}^2 = \sigma_t^2 - \alpha_{t \mid s}^2 \sigma_s^2$

The scheduling parameters, $\alpha_t$ and $\sigma_t$ are strictly positive, smooth, monotonically decreasing and increasing functions of time respectively. Based on this we can define the Signal-to-Noise ratio $\text{SNR(t)}$ as:

\begin{equation}
\text{SNR}(t) = \frac{\alpha_t^2}{\sigma_t^2}
\end{equation}
 
As time $t$ progresses, the \text{SNR} decreases. This implies that for $s < t$, we have $\text{SNR}(s) > \text{SNR}(t)$.
At t=0 the data is least noisy and at t=1 there is no more signal left in the data, hence $q(\rvz_1 \mid \rvx) = \mathcal{N}(\rvz_1; 0,\mI)$.

The choice of schedule significantly impacts the performance of diffusion models, and there are several ways of noise scheduling. 
DDPM \cite{ho2020denoising} employed a linear schedule to add noise over 1000 discrete timesteps. Nichol and Dhariwal \cite{nichol2021improved}, used cosine scheduling and found it to perform better due to its smooth transition between low and high levels of noise.
In VDM \cite{kingma2021variational}, the authors learned the forward noise schedule, moreover they demonstrate that increasing the number of timesteps resulted in a decrease in loss, thereby achieving good results with a continuous-time model.
The schedules used in the above mentioned works were variance preserving ($\alpha_t^2 = 1 - \sigma_t^2$), which ensures that the variance of the data remains constant throughout the forward process.
Alternatively, in the case of variance-exploding schedules \cite{song2020score,song2020improved}, $\alpha_t^2 = 1$.
It was demonstrated by Kingma et al. \cite{kingma2021variational} that the variance preserving and variance exploding formulations can be considered equivalent in continuous time.

\subsection{Reverse generative process}
\label{subsec:ReverseDiff}

The reverse diffusion process $q(\rvz_{s} \mid \rvz_t)$ is also a Markov chain with Gaussian transition probability and aims to recover the original data $\rvx$ from the noisy data $\rvz_t$. 
Since the true reverse process $q(\rvz_{s} \mid \rvz_t)$ is intractable, it is approximated with a learned distribution $p_\vtheta(\rvz_s \mid \rvz_t)$. This forms a hierarchical generative model that samples a sequence of latent variables $\rvz_t$, with time progressing from t=1 to t=0, gradually denoising the data over $T$ steps to recover the original distribution. For discrete time case number of steps $T$ is finite and is discretized into uniform timesteps of width $1/T$, with $s(i) = \frac{i - 1}{T}$ and $t(i) = \frac{i}{T}$, 

The overall reverse process is defined as,

\begin{equation}
p_\vtheta(\rvx) = \int_{\rvz} p(\rvz_1) p_\vtheta(\rvx \mid \rvz_0) \prod_{i=1}^T p_\vtheta(\rvz_{s(i)} \mid \rvz_{t(i)})d\rvz
\end{equation}

To approximate the true data distribution we need to minimize the negative log likelihood. 
However, that is intractable and we minimize the tractable negative variational lower bound also called negative evidence lower bound (NELBO) instead, which is standard in latent variable models and is expressed as, 

\begin{equation}
\begin{aligned}
- \log p_\vtheta(\rvx) &\leq \text{NELBO}(\rvx) = 
\underbrace{D_\text{KL}\left(q(\rvz_1 \mid \rvx) \, \| \, p(\rvz_1)\right)}_{\text{Prior Loss}} + \\
& \underbrace{\mathbb{E}_{q(\rvz_0 \mid \rvx)}\left[- \log p_\vtheta(\rvx \mid \rvz_0)\right]}_{\text{Reconstruction Loss}} + \\
& \underbrace{\sum_{i=1}^T \mathbb{E}_{q(\rvz_{t(i)} \mid \rvx)} \, D_\text{KL}\left[q(\rvz_{s(i)} \mid \rvz_{t(i)}, \rvx) \, \| \, p_\vtheta(\rvz_{s(i)} \mid \rvz_{t(i)})\right]}_{\text{Diffusion Loss } (L_T(\rvx))}
\end{aligned}
\label{revprocesselbo}
\end{equation}

Based on the assumptions of the forward process, $\rvz_0$ is nearly identical to $\rvx$ because only a small amount of noise is added, making the reconstruction loss in the equation \eqref{revprocesselbo} negligible and therefore can be dropped from the objective in practice.
Moreover, as discussed in section \ref{subsec:ForwardDiff}, $q(\rvz_1 \mid \rvx)$ approaches a pure Gaussian distribution at the end of the forward process which matches our fixed prior $p(\rvz_1) = \mathcal{N}(\rvz_1; 0,\mI)$. 
As a result, the KL divergence $D_\text{KL}\big(q(\rvz_1 \mid \rvx) \,\|\, p(\rvz_1)\big)$ tends to zero, hence this term is also dropped.
The remaining term is the diffusion loss $L_T(\rvx)$ which depends on the number of timesteps $T$ determining the depth of the generative model.

\section{Loss formulations}
\label{Loss_forms}

In the previous section, we defined the NELBO objective \eqref{revprocesselbo}. 
For the denoising model there are several options for the target prediction in addition to the data $\rvx$. 
For example, some approaches focus on predicting the noise $\rvepsilon$ added during the forward process \cite{ho2020denoising,song2020denoising,nichol2021improved}. Another approach predicts the rate of change in the data distribution over time, also known as $\rvv$-prediction \cite{salimans2022progressive}.
Some methods target the score function $\nabla_{\rvx} \log p(\rvx)$ \cite{song2019generative,song2020score}, which is the gradient of the log-probability density of the data.

For each of these targets, we can derive the NELBO loss formulation ($L$) from equation \eqref{revprocesselbo}. 
In addition, other loss formulations are also proposed in the literature, typically designed to prioritize perceptual sample quality or computational efficiency. 
We call these weighted loss functions ($\mathcal{L}$) as they can all be shown as a weighted function of the NELBO where the weight $w(t)$ is a suitable chosen weighting function. 

\begin{equation}
\mathcal{L} = w(t)L
\end{equation}

In the following sections, we explore the various target predictions and corresponding loss formulations in detail. 
We present a systematic review of these relationships, unifying them under the framework of the NELBO objective. Specifically, we derive the NELBO in terms of these alternative targets and show that all the different objectives, whether predicting the original data $\rvx$, noise $\rvepsilon$, rate of change in the data distribution $\rvv$, or score $\rvs$ can be expressed as weighted functions of the NELBO. 
For clarity, we refer to different target objectives as $\rvx$-space, $\rvepsilon$-space, $\rvv$-space, and $\rvs$-space throughout this paper.

\subsection{$\rvx$-space}

As shown in section \ref{subsec:ReverseDiff}, the NELBO reduces to diffusion loss which is the last term of equation \eqref{revprocesselbo}. This can be further simplified to the following form (a detailed derivation of these steps is provided in appendix B.1).

\begin{equation}
\begin{aligned}
L_T(\rvx) &= \frac{T}{2} \, \mathbb{E}_{\rvepsilon \sim \mathcal{N}(\vzero, \mI), i \sim \text{U}\{1, T\}} 
\big[(\text{SNR}(s(i)) - \text{SNR}(t(i))) \, \| \rvx - \hat{\rvx}_\vtheta(\rvz_t; t) \|_2^2 \big]
\end{aligned}
\label{finite_t_loss}
\end{equation}

where $\hat{\rvx}_\vtheta(\rvz_t; t)$ is the prediction of the original data $\rvx$ by our denoising model given the noisy data $\rvz_t = \alpha_t\rvx + \sigma_t\rvepsilon$ at timestep $t$.

For the continuous-time case, $T \to \infty$. 
Here, the timestep $t$ is treated as a continuous variable, and the transition process is referred to as the continuous-time diffusion process \cite{kingma2021variational}. In this setting, equation \eqref{finite_t_loss} transforms into the following form,

\begin{equation}
\begin{aligned}
L(\rvx) &= - \mathbb{E}_{t \sim \mathcal{U}(0, 1), \rvepsilon \sim \mathcal{N}(\vzero, \mI)} \big[\text{SNR}'(t) \, \| \rvx - \hat{\rvx}_\vtheta(\rvz_t; t) \|_2^2 \big]
\end{aligned}
\label{infinite_t_loss}
\end{equation}

where we prove that for cosine noise schedule, $\text{SNR}'(t) = \frac{-\pi\alpha_t}{\sigma_t^3}$ (see appendix B.1). Note that we use $\text{U}\{1, T\}$ to denote sampling from a discrete uniform distribution, while $\mathcal{U}(0, 1)$ denotes sampling from a continuous uniform distribution in the continuous-time setting.

Moreover we can define the weighted loss as,

\begin{equation}
\begin{aligned}
\mathcal{L}(\rvx) &= - \mathbb{E}_{t \sim \mathcal{U}(0, 1), \rvepsilon \sim \mathcal{N}(\vzero, \mI)} \big[ w_{\rvx}(t) \, \text{SNR}'(t) \, \| \rvx - \hat{\rvx}_\vtheta(\rvz_t; t) \|_2^2 \big]
\end{aligned}
\label{wt_x_loss_1}
\end{equation}

By choosing $w_{\rvx}(t) = -\frac{1}{\text{SNR}'(t)}$, this further simplifies as an expected value of the mean squared error between the original data and the predicted data,

\begin{equation}
\begin{aligned}
\mathcal{L}(\rvx) &= \mathbb{E}_{t \sim \mathcal{U}(0, 1), \rvepsilon \sim \mathcal{N}(\vzero, \mI)} \left[\| \rvx -\hat{\rvx}_\vtheta(\rvz_t; t) \|_2^2 \right] = w_{\rvx}(t) L(\rvx)
\end{aligned}
\label{weighted_x_loss}
\end{equation}

\subsection{$\rvepsilon$-space}
The $\rvepsilon$-space loss formulation is one of the most commonly used objective in diffusion models, as proposed in DDPM\cite{ho2020denoising}. Instead of directly reconstructing the original data $\rvx$, we model the noise component $\rvepsilon$ that was added to the data in every time step during the forward diffusion process. 
The authors of the paper claimed that this approach simplifies the learning task, as the prediction of noise aligns with the stochastic nature of the diffusion process.

We derive the NELBO loss in $\rvepsilon$-space, as detailed in the appendix B.2,

\begin{equation}
\begin{aligned}
L(\rvepsilon) &= - {\mathbb{E}}_{t, \epsilon} 
\left[ \frac{\text{SNR}'(t)}{\text{SNR}(t)} \| \rvepsilon - \hat{\rvepsilon}_\vtheta(\rvz_t; t) \|_2^2 \right]
\end{aligned}
\label{elbo_epsilon_loss}
\end{equation}

The loss proposed in DDPM is different from the NELBO loss. 
They used the weighted $\rvepsilon$ loss, which implies that the model learns to predict the noise sampled from the unit Gaussian and not the scaled noise which was added to the original data $\rvx$ at every timestep during the forward diffusion process. 
The weighted $\rvepsilon$-loss is given as below and can be seen as a weighted function of \eqref{elbo_epsilon_loss} with weight $w_{\rvepsilon}(t) = -\frac{\text{SNR}(t)}{\text{SNR}'(t)}$:

\begin{equation}
\begin{aligned}
\mathcal{L}(\rvepsilon) &= {\mathbb{E}}_{t, \epsilon}
\big[\| \rvepsilon - \hat{\rvepsilon}_\vtheta(\rvz_t; t)\|_2^2 \big] = w_{\rvepsilon}(t) L(\rvepsilon)
\end{aligned}
\label{weighted_epsilon_loss}
\end{equation}

\subsection{v-space}
\label{vspace}
The v-space loss, introduced in \cite{salimans2022progressive}, combines the data $\rvx$ and noise $\rvepsilon$. 
This formulation is particularly beneficial for model distillation to reduce the number of sampling steps, as while sampling the standard noise objective becomes unstable when the SNR approaches zero. 
In such cases $\alpha_t$ tends to zero, leading to instability in reconstructing the data from the predicted noise as $\hat{\rvx}_\vtheta(\rvz_t) = \frac{1}{\alpha_t} \left( \rvz_t - \sigma_t \hat{\boldsymbol{\epsilon}}_\vtheta(\rvz_t) \right)$. 
The authors showed that this issue has less impact in conventional diffusion models, where clipping the reconstructed data in the desired range and using a large number of sampling steps can mitigate errors but becomes a key factor in distillation, where efficient sampling is essential in a limited number of steps.

This approach expresses the noisy data using an angular parameter $\phi_t$, where $\rvz_{\phi_t} = \cos(\phi_t) \rvx + \sin(\phi_t) \rvepsilon$, and $\phi_t = \arctan\left( \frac{\sigma_t}{\alpha_t} \right)$. 
The target $\rvv_{\phi_t}$ is then calculated as $\rvv_{\phi_t} = \frac{d\rvz_{\phi_t}}{d\phi_t} = \cos(\phi_t) \rvepsilon - \sin(\phi_t) \rvx$, which represents the instantaneous direction and rate of change required to transform the noisy data $\rvz_{\phi_t}$ along a circular trajectory parameterized by the angle $\phi_t$. 

The NELBO loss in $\rvv$-space as derived in appendix B.3 is given as,

\begin{equation}
\begin{aligned}
L(\rvv) = -{\mathbb{E}}_{t, \epsilon}
\left[ \frac{\sigma_t^2}{\alpha_t^2 + \sigma_t^2} \, \text{SNR}'(t) \, \| \rvv - \hat{\rvv}_\vtheta(\rvz_t; t) \|_2^2 \right]
\end{aligned}
\label{elbo_v_loss}
\end{equation}

The weighted $\rvv$ loss can be formulated as the weighted function of NELBO with weight $w_{\rvv}(t) = -\frac{(\alpha_t^2 + \sigma_t^2)}{\sigma_t^2 \text{SNR}'(t)}$

\begin{equation}
\begin{aligned}
\mathcal{L}(\rvv) &= {\mathbb{E}}_{t, \epsilon}
\big[\| \rvv - \hat{\rvv}_\vtheta(\rvz_t; t)\|_2^2 \big] = w_{\rvv}(t) L(\rvv)
\end{aligned}
\label{weighted_v_loss}
\end{equation}

\subsection{$\rvs$-space}

Score modeling, introduced by Song et al. \cite {song2019generative} uses denoising score matching \cite{vincent2011connection} to approximate the score function and then use a neural network to learn it. The idea behind score matching is to add a small amount of noise to the data, which makes the score calculation tractable, and therefore learn the score of perturbed distribution  instead of the original distribution, which is expressed as $\nabla_{\rvz_t} \log q(\rvz_t \mid \rvx)$. 
The authors demonstrate that minimizing the denoising score matching objective across multiple noise scales enables high quality sample generation. 
This theory is closely aligned with the diffusion process, as both approaches aim to refine the noisy data toward its original distribution.

In \cite{song2020score}, the authors bridged the gap between score modeling and diffusion models by proposing score based modeling using stochastic differential equations (SDE). 
They showed that the forward diffusion process can be interpreted as a discretization of a continuous-time SDE, and the reverse process corresponds to solving the reverse-time SDE using the learned score function.

We demonstrate here that the NELBO loss can once again be formulated in $\rvs$-space and can be expressed as below (see in appendix B.4),

\begin{equation}
\begin{aligned}
L(\rvs) = - {\mathbb{E}}_{t, \epsilon}
\left[ \frac{\sigma_t^4}{\alpha_t^2} \text{SNR}'(t) \, \| \nabla_{\rvz_t} \log q(\rvz_t \mid \rvx) - \hat{\rvs}_\vtheta(\rvz_t; t) \|_2^2 \right]
\end{aligned}
\label{elbo_score_loss}
\end{equation}

The weighted loss in $\rvs$-space can be formulated as the weighted function of NELBO with weight $w_{\rvs}(t) = -\frac{\alpha_t^2}{\sigma_t^4 \text{SNR}'(t)}$

\begin{equation}
\begin{aligned}
\mathcal{L}(\rvs) = {\mathbb{E}}_{t, \epsilon}
\big[\| \nabla_{\rvz_t} \log q(\rvz_t \mid \rvx) - \hat{\rvs}_\vtheta(\rvz_t; t) \|_2^2 \big] = w_{\rvs}(t) L(\rvs)
\end{aligned}
\label{weighted_score_loss}
\end{equation}

Furthermore, for Gaussian noise perturbation the score simplifies to:

\begin{equation}
\begin{aligned}
\rvs = \nabla_{\rvz_t} \log q(\rvz_t \mid \rvx) = -\frac{(\rvz_t - \alpha_t \rvx)}{\sigma_t^2}
\end{aligned}
\label{score_gaussian}
\end{equation}

\subsection{Equivalence of loss functions}
\label{Loss_eq}

All the NELBO loss formulations across different parameter spaces are derived from same equation \eqref{infinite_t_loss}, therefore, they are fundamentally equivalent. 
However, to derive the weighted loss from the NELBO loss, we need to apply different weights, which essentially removes the SNR scalings associated with the $\ell_2$ difference between the target and the prediction. 
As a result, these weighted loss functions are not equivalent, even though they are all initially derived from the same NELBO formulation. 

To establish a relationship between the weighted loss formulations, we rescale the weighted loss in $\rvepsilon$, $\rvv$ and $\rvs$ space to make it equal to $\mathcal{L}(\rvx)$. We call them rescaled loss $\widetilde{\mathcal{L}}$ that is equivalent across all targets and can be easily obtained using the weights derived in the previous sections,

\begin{equation}
\begin{aligned}
\widetilde{\mathcal{L}}(\rvepsilon) = \frac{\sigma_t^2}{\alpha_t^2}\mathcal{L}(\rvepsilon) = \mathcal{L}(\rvx)
\end{aligned}
\label{epsilon_eq_x}
\end{equation}

\begin{equation}
\begin{aligned}
\widetilde{\mathcal{L}}(\rvv) = \frac{\sigma_t^2}{\alpha_t^2 + \sigma_t^2}\mathcal{L}(\rvv) = \mathcal{L}(\rvx)
\end{aligned}
\label{v_eq_x}
\end{equation}

\begin{equation}
\begin{aligned}
\widetilde{\mathcal{L}}(\rvs) = \frac{\sigma_t^4}{\alpha_t^2}\mathcal{L}(\rvs) = \mathcal{L}(\rvx)
\end{aligned}
\label{score_eq_x}
\end{equation}

Table \ref{tab:loss_overview} summarizes all the loss formulations across different targets for the denoising model.

\begin{table}[h]
    \centering
    \caption{Overview of all the loss formulations across different scenarios. While the NELBO and the rescaled loss are equivalent and comparable, the weighted losses are not equivalent and are expected to exhibit different empirical performance.}
    \begin{tabular}{|p{1.2cm}|p{4.7cm}|p{2.7cm}|p{3cm}|}
        \hline
        \centering {\textbf{Target}} & \centering {\textbf{NELBO loss} \\($L$)} & \centering {\textbf{Weighted loss}\\($\mathcal{L}$)}& \centering {\textbf{Rescaled loss}\\ ($\widetilde{\mathcal{L}}$)}\tabularnewline
        \hline
        \centering $\rvx$ & \centering $-\mathbb{E} \big[\text{SNR}'(t) \, \| \rvx - \hat{\rvx}_\vtheta \|_2^2 \big]$ & \centering $\mathbb{E}\big[\| \rvx - \hat{\rvx}_\vtheta \|_2^2 \big]$ & \centering $\mathbb{E}\big[\| \rvx - \hat{\rvx}_\vtheta \|_2^2 \big]$ \tabularnewline
        \hline
        \centering$\rvepsilon$ & \centering $-\mathbb{E}\big[ \frac{\text{SNR}'(t)}{\text{SNR}(t)} \| \rvepsilon - \hat{\rvepsilon}_\vtheta \|_2^2 \big]$ & \centering $\mathbb{E}\big[\| \rvepsilon - \hat{\rvepsilon}_\vtheta \|_2^2 \big]$ & \centering $\mathbb{E}\big[\frac{\sigma_t^2}{\alpha_t^2} \| \rvepsilon - \hat{\rvepsilon}_\vtheta \|_2^2\big]$ \tabularnewline
        \hline
        \centering$\rvv$ & \centering $- \mathbb{E}\big[ \frac{\sigma_t^2}{\alpha_t^2 + \sigma_t^2} \, \text{SNR}'(t) \, \| \rvv - \hat{\rvv}_\vtheta \|_2^2 \big]$ & \centering $\mathbb{E}\big[\| \rvv - \hat{\rvv}_\vtheta \|_2^2 \big]$ & \centering $\mathbb{E}\big[\frac{\sigma_t^2}{\alpha_t^2 + \sigma_t^2} \| \rvv - \hat{\rvv}_\vtheta\|_2^2\big]$ \tabularnewline
        \hline
        \centering $\rvs$ & \centering $- \mathbb{E}\big[ \frac{\sigma_t^4}{\alpha_t^2} \text{SNR}'(t) \, \| \rvs - \hat{\rvs}_\vtheta \|_2^2 \big]$ & \centering {$\mathbb{E}\big[\| \rvs - \hat{\rvs}_\vtheta \|_2^2 \big]$} & \centering $\mathbb{E}\big[\frac{\sigma_t^4}{\alpha_t^2} \| \rvs - \hat{\rvs}_\vtheta\|_2^2\big]$ \tabularnewline
        \hline
    \end{tabular}
\label{tab:loss_overview}
\end{table}

Although researchers have claimed that some training objectives outperform others in diffusion models, the reasons behind these differences remain unclear and are often attributed to empirical observations rather than theoretical foundations. 
For instance, while some works prefer more complex weights for loss functions \cite{choi2022perception,karras2022elucidating,hang2023efficient}, others \cite{ho2020denoising,nichol2021improved} find that simpler objectives (e.g. $\ell_2$ loss between target and prediction) perform just as well or even better in practice. 
This discrepancy raises questions about the fundamental role of loss formulations in training diffusion models and whether the observed performance gaps are due to the loss functions themselves or other factors such as model architecture, training dynamics, or noise schedules.

In our theoretical analysis, we formulated the NELBO loss for different denoising models. 
Specifically, we showed that different formulations of the learned model (i.e. predicting original data $\rvx$, noise $\rvepsilon$, rate of change of data distribution $\rvv$ and score function $\rvs$) can be mapped to one another, and their corresponding NELBO objectives are mathematically interchangeable. 
We also formulated the relations between the weighted loss formulations. 

In principle, the mathematical equivalence we established should hold when we train the various denoising models with equivalent loss formulations under similar conditions (e.g., dataset, model architecture etc.).
In the next section, we outline the experiments conducted to validate this hypothesis and provide a detailed analysis of the results obtained.

\section{Experiments}
\label{sec:Experiments}
In this section, we outline the experimental setup used to conduct our tests and present the results obtained from these experiments. Additionally, we give a detailed analysis and insights into the findings.

\subsection{Experimental setup}

\begin{figure}[h!]
    \centering
    \includegraphics[width=12cm, height = 3cm]{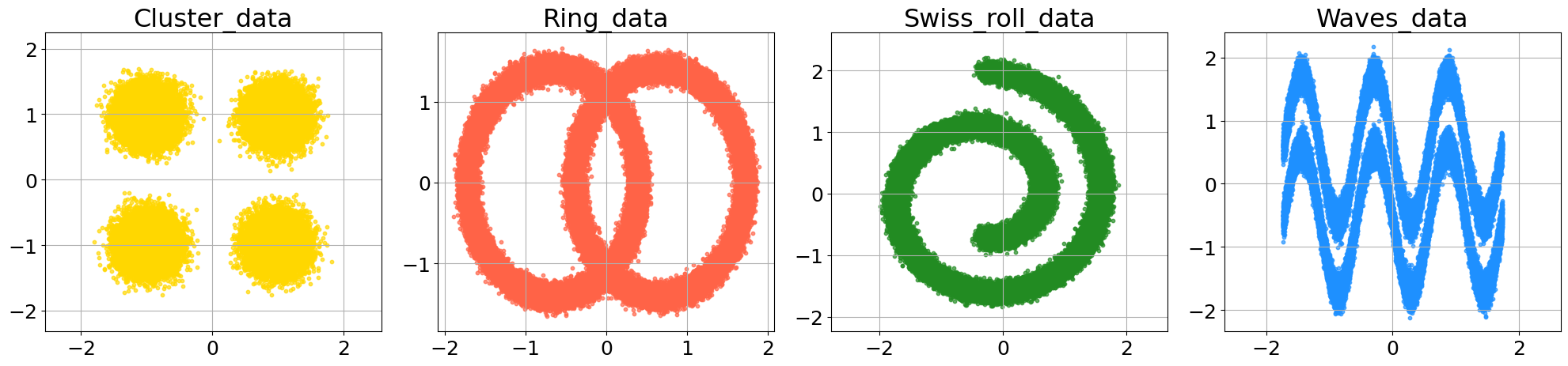}
    \caption{Scatter plot for 2D datasets}
    \label{2D_data}
\end{figure}

\begin{figure}[h!]
    \centering
    \includegraphics[width=11.5cm, height = 5cm]{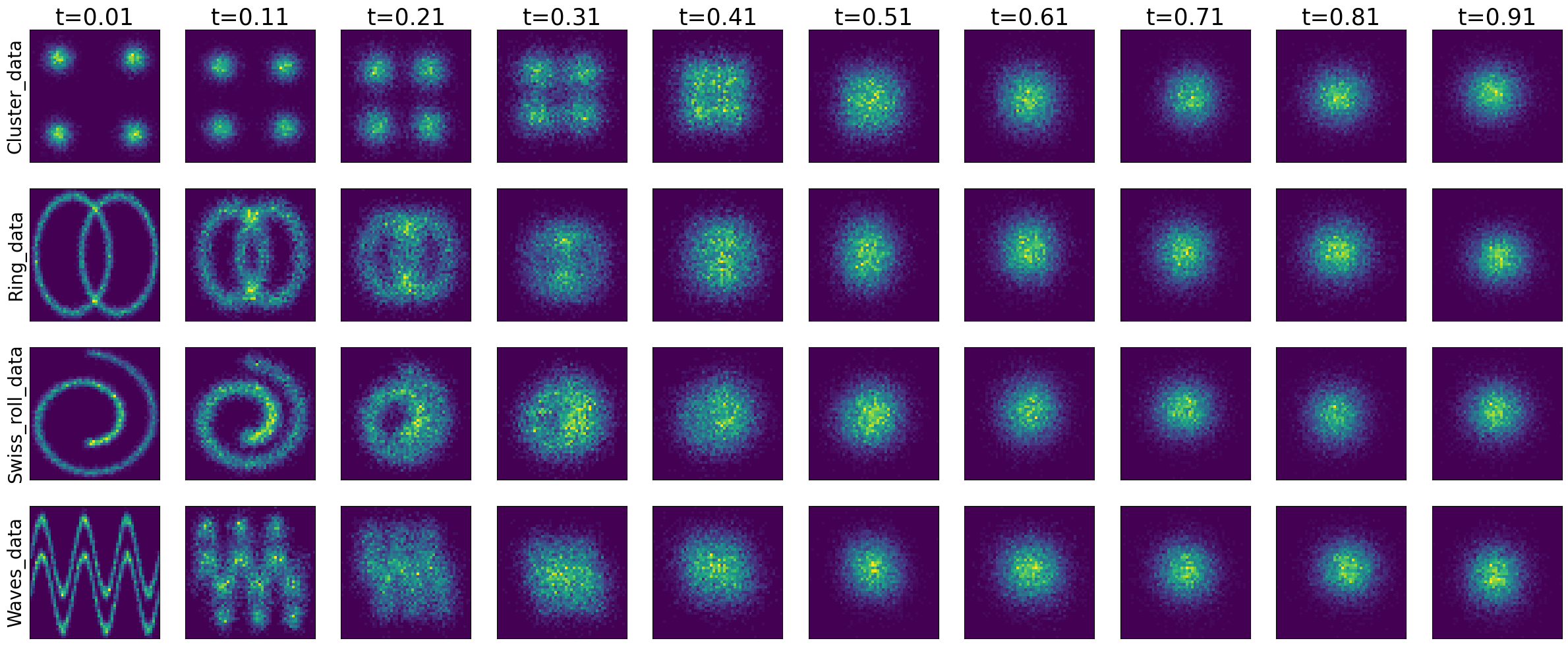}
    \caption{Effects of adding cosine scheduled Gaussian noise in the forward process}
    \label{2D_data_noise}
\end{figure}

To conduct our experiments, we first work with 2-dimensional synthetic datasets that we generated ourselves. These datasets are well-suited for detailed analysis as it is easy to plot numerous examples and visually analyze the complete data manifold. To ensure the generalizability of our findings, we select four distinct 2D datasets with 100K samples each. These datasets are: Cluster data, Ring data, Swiss roll data and Waves data, the scatter plots of these datasets can be seen in fig. \ref{2D_data}. In fig. \ref{2D_data_noise} we show the effect of gaussian noise added in the forward process for all these datasets. 

We also perform experiments on a high-dimensional image dataset, CIFAR-10, which is a publicly available dataset that contains 32x32 color images across 10 classes. While we present results on an image dataset, our main focus is not on extensive image generation experiments but understanding the behavior of different loss formulations. However, this work sets the foundation for future research to explore their impact on image data more deeply.

We used a variance-preserving cosine schedule for the forward process, combined with a continuous-time reverse model $T \to \infty$. Hence, the noisy data at time $t \sim \mathcal{U}(0, 1)$, is given as $\rvz_t = \cos(0.5\pi t)\rvx + \sin(0.5\pi t)\rvepsilon$. To ensure comparability across experiments, for the 2D datasets, we modeled the reverse process using a simple feedforward neural network architecture consisting of 7 fully connected layers followed by a ReLU activation. In addition, we maintained consistent training dynamics for all datasets. For the image dataset, we used an architecture inspired by diffusers UNet model \cite{ronneberger2015u}.

\subsection{Experimental results and discussion}

In our analysis, we examine the performance of diffusion model trained with various loss formulations from three key perspectives: $(\text{i})$ loss convergence over epochs indicating the training and stability efficiency, $(\text{ii})$ the quality of generated samples that reveals how well the model produces realistic and high fidelity samples, and $(\text{iii})$ loss behavior at different timesteps $t$ that give insights into how different loss formulations influence the reverse diffusion process over time. Due to space limitations, we present some results only for the ring data, while results for other datasets follow similar patterns and are provided in the appendix C for completeness.

\subsubsection{Loss convergence vs epochs:}
\label{lossvsEpochs}

\begin{figure}[h!]
    \includegraphics[width=12cm, height = 3cm]{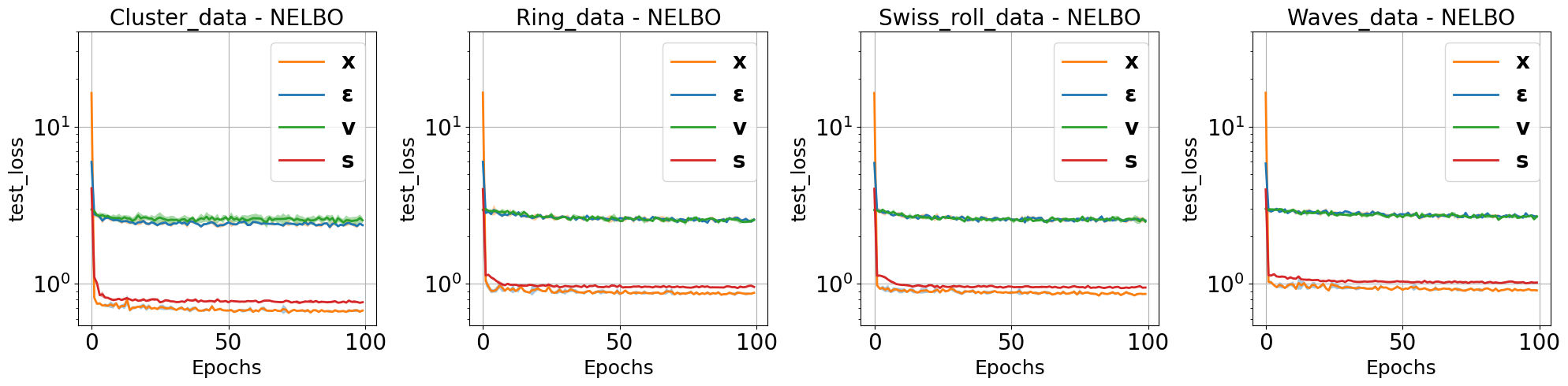}
    \caption{NELBO test loss for different datasets}
    \label{Elbo_loss_curve}
\end{figure}

We begin by training the denoising model using the NELBO loss formulations $L$ for different target predictions. Given their theoretical equivalence as discussed in section \ref{Loss_forms}, we expect them to behave similarly in experiments.
Fig. \ref{Elbo_loss_curve} illustrates the NELBO test loss for different datasets. The loss curves for predictions in the $\rvv$ and $\rvepsilon$ space are close, and so is loss in $\rvx$ and $\rvs$ space. However, these two groups differ significantly for all scenarios indicating a discrepancy in their training dynamics. 

\begin{figure}[h!]
    \centering
    \includegraphics[width=12cm, height = 3cm]{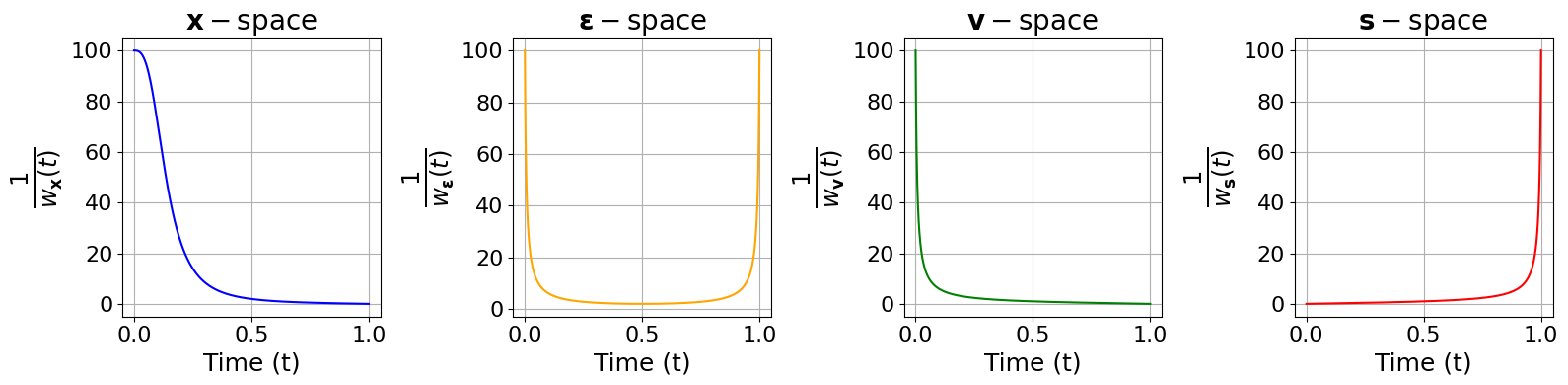}
    \caption{$\text{SNR}$ scalings ($\frac{1}{w(t)}$) with respect to timesteps for various NELBO formulations}
    \label{ELBO_coeff}
\end{figure}

We attribute these differences to the different $\text{SNR}$ scaling of the targets within the NELBO formulation which is inversely proportional to the weighting function and is given by $\frac{1}{w(t)}$. These scaling factors control how much each timestep contributes to the overall loss, and therefore have an impact on how the model learns during training.
As shown in fig. \ref{ELBO_coeff}, the scaling for $\rvepsilon$ and $\rvv$ space are substantially higher in the early time steps, when the noise added to the data is minimal. While $\rvx$ space also exhibits large initial scaling, its decay is more gradual over time. In contrast, the $\rvs$ space has higher scaling at later timesteps. This pattern suggests that excessively high scaling at early timesteps, when noise levels are low, may negatively impact the model's overall likelihood performance.

\begin{figure}[h!]
    \centering
    \includegraphics[width= 12cm, height = 5cm]{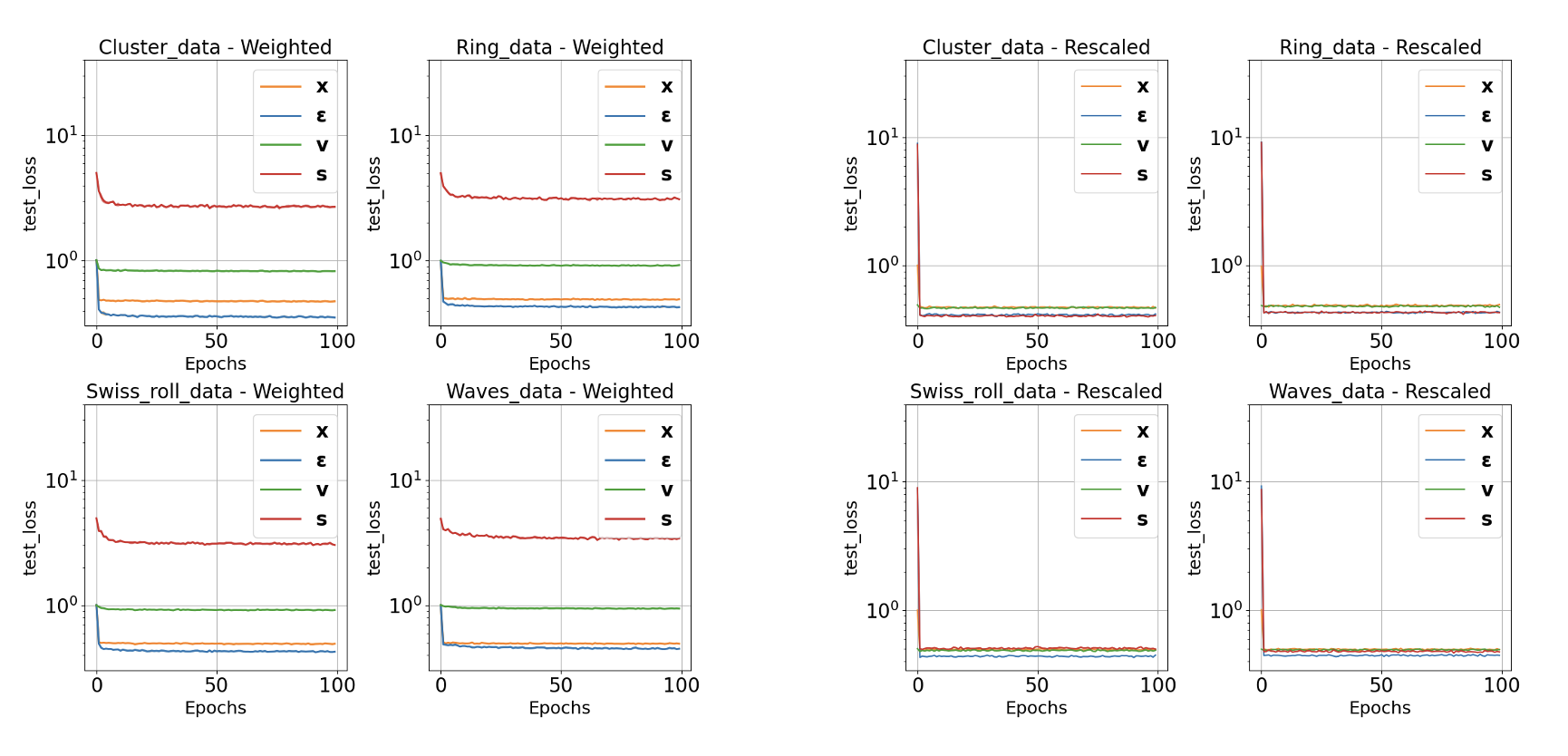}
    \caption{The weighted test loss $\mathcal{L}$ (left), is not directly comparable across different target predictions. However, the rescaled test loss $\tilde{\mathcal{L}}$ (right), is comparable and demonstrates the mathematical equivalence discussed in section \ref{Loss_eq}.}
    \label{Weighted_x_rescaled}
\end{figure}

Next, we train the model using the weighted loss formulations $\mathcal{L}$ for different datasets as shown in fig. \ref{Weighted_x_rescaled} (left). As outlined in section \ref{Loss_eq}, the weighted loss formulations are not equivalent and therefore not comparable. To address this, we rescale the weighted test loss as defined in equations \eqref{epsilon_eq_x}, \eqref{v_eq_x}, and \eqref{score_eq_x}. This gives the rescaled loss, $\tilde{\mathcal{L}}$, which is mathematically equivalent to $\mathcal{L}(\rvx)$. The rescaled loss is plotted in fig. \ref{Weighted_x_rescaled} (right), where we observe that, after rescaling, the loss curves are very close to each other. This confirms that the mathematical equivalency holds. Moreover, this indicates that the weighted loss formulation is more stable compared to the NELBO formulations, as there are no additional factors influencing the training dynamics.

\subsubsection{Generated samples:}
\begin{figure}[h!]
    \centering
    \includegraphics[width=8cm, height = 3.8cm]{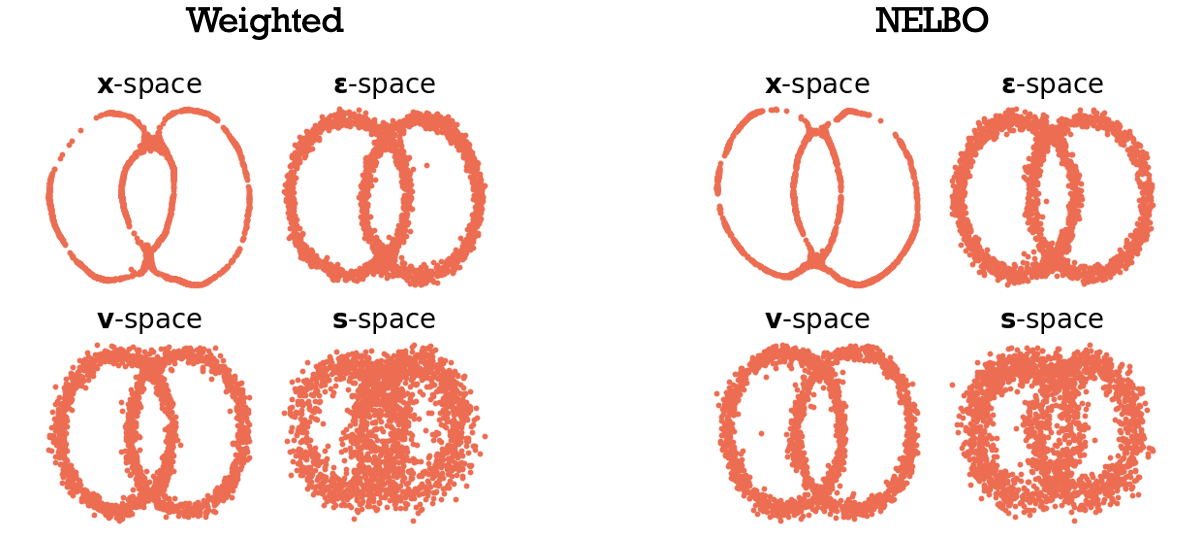}
    \caption{Comparing 2K samples generated after 512 sampling steps from model trained using weighted loss formulation (left) and NELBO formulation (right)}
    \label{Samples_both}
\end{figure}

The quality of generated samples shows a different trend compared to loss convergence, indicating that better likelihood estimation does not necessarily correlate with better sample generation, as also discussed in \cite{theis2015note}.
To analyze the discrepancy, we compare the sample quality using moment-based metrics. Specifically, we measure the mean distance (Euclidean distance between dataset means) and covariance distance (Frobenius norm of the difference between covariance matrices) between real and generated samples.
The results are shown in table \ref{tab:metrics_comparison}. We see that although the NELBO is better for $\rvx$ and $\rvs$ space the sample quality is better for $\rvepsilon$ and $\rvv$ space. Moreover, the sample quality for weighted and NELBO loss are similar in most of the cases as also illustrated in fig. \ref{Samples_both} which shows 2K generated samples for the ring dataset using weighted and NELBO loss formulations, respectively.
This suggests that while the scaling in the NELBO loss functions influences how the model converges, it has little effect on the quality of the generated samples.

\begin{table}[h!]
\centering
\caption{Comparison of NELBO and weighted loss formulations for 2D datasets}
\begin{tabular}{|p{1.8cm}|p{0.8cm}|p{1.1cm}|p{1cm}|p{1cm}|p{1cm}|p{1cm}|p{1cm}|}
\hline
\centering{\textbf{Data}} & \centering{\textbf{Loss Form}} & \multicolumn{3}{c|}{\textbf{\makecell{NELBO Loss \\ ($L$)}}} & \multicolumn{3}{c|}{\textbf{\makecell{Weighted loss \\ ($\mathcal{L}$)}}} \\ \cline{3-8}
& & \textbf{Nelbo$\downarrow$} & \textbf{Mean dist.$\downarrow$} & \textbf{Covar dist.$\downarrow$} & \textbf{Loss$\downarrow$} & \textbf{Mean dist.$\downarrow$} & \textbf{Covar dist.$\downarrow$} \\ \hline
\centering{Cluster data} & \centering{$\rvx$} & 0.6777 & 0.1754 & 0.5746 & 0.4754 & 0.4300 & 0.2715\\ 
& \centering{$\rvepsilon$} & 2.3636 & 0.0364 & 0.0706 & 0.3522 & 0.0634 & 0.1430\\ 
& \centering{$\rvv$} & 2.5396 & 0.0307 & 0.0409 & 0.8264 & 0.0389 & 0.0363\\ 
& \centering{$\rvs$} & 0.7657 & 0.2279 & 0.1498 & 2.6934 & 0.2688 & 0.1633\\ \hline
\centering{Ring data} & \centering{$\rvx$} & 0.8785 & 0.2744 & 0.3807 & 0.4932 & 0.2807 & 0.3974\\ 
& \centering{$\rvepsilon$} & 2.5700 & 0.0914 & 0.1107 & 0.4266 & 0.0983 & 0.0366\\ 
& \centering{$\rvv$} & 2.5452 & 0.0459 & 0.0718 & 0.9227 & 0.0453 & 0.0088\\ 
& \centering{$\rvs$} & 0.9577 & 0.2254 & 0.1563 & 3.0981 & 0.2220 & 0.1637\\ \hline
\centering{Swiss data} & \centering{$\rvx$} & 0.8640 & 0.1133 & 0.2645 & 0.4934 & 0.5256 & 0.6875\\ 
& \centering{$\rvepsilon$} & 2.5324 & 0.0689 & 0.0941 & 0.4261 & 0.0857 & 0.0824\\ 
& \centering{$\rvv$} & 2.4861 & 0.0418 & 0.0598 & 0.9171 & 0.0427 & 0.0269\\ 
& \centering{$\rvs$} & 0.9493 & 0.1266 & 0.1972 & 3.0274 & 0.0893 & 0.1227\\ \hline
\centering{Waves data} & \centering{$\rvx$} & 0.9104 & 0.1593 & 0.5559 & 0.4939 & 0.1869 & 0.6911\\ 
& \centering{$\rvepsilon$} & 2.6805 & 0.0405 & 0.0757 & 0.4500 & 0.0748 & 0.0778\\ 
& \centering{$\rvv$} & 2.6873 & 0.0447 & 0.0738 & 0.9411 & 0.0131 & 0.0271\\ 
& \centering{$\rvs$} & 1.0210 & 0.0353 & 0.1369 & 3.4165 & 0.0178 & 0.1676\\ \hline
\end{tabular}
\label{tab:metrics_comparison}
\end{table}

\subsubsection{Loss vs timesteps:} 
\begin{figure}[h!]
    \centering
    \includegraphics[width=10cm, height = 5cm]{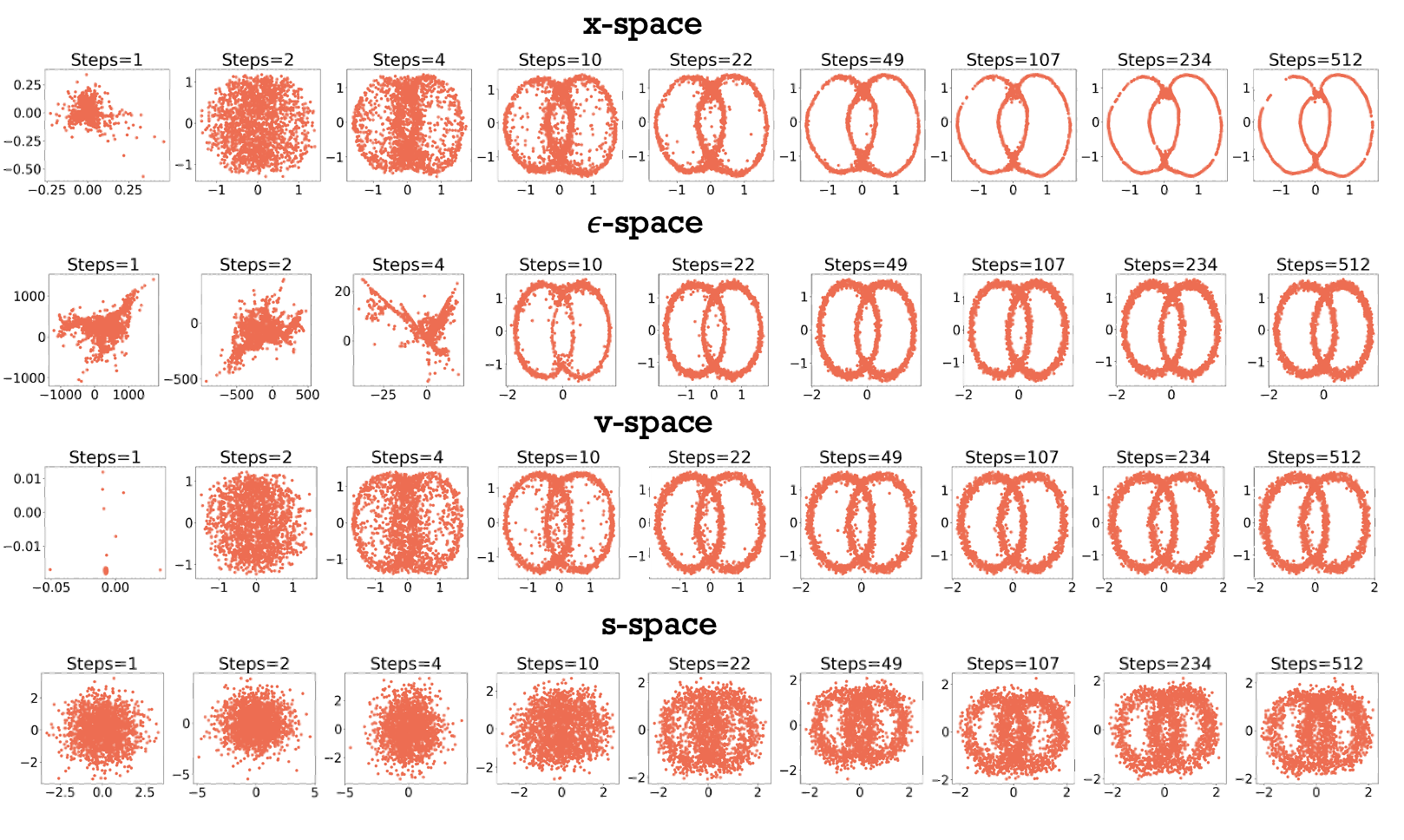}
    \caption{Generated samples for ring data for different number of sampling steps from model trained on weighted loss $\mathcal{L}$}
    \label{Ring Sampling Steps}
\end{figure}

In fig. \ref{Ring Sampling Steps}, we illustrate the generation of samples using different numbers of sampling steps in the reverse process for the model trained with the weighted loss. The results are similar to those observed with the NELBO loss (see appendix C.1). It can be seen in the image that for the $\rvx$-space, sample quality declines with more sampling steps but outperforms other objectives with fewer steps, effectively capturing data structure and scale. In contrast, the $\rvepsilon$-space produces poorer samples with fewer steps, and the sample quality gradually increases. The $\rvv$-space, captures the data structure well even with fewer sampling steps and the sample quality continues to improve with more steps. The quality of samples generated in the $\rvs$-space is not good, however, it improves with the number of steps. 

\begin{figure}[h!]
    \centering
    \includegraphics[width=11cm, height = 2.5cm]{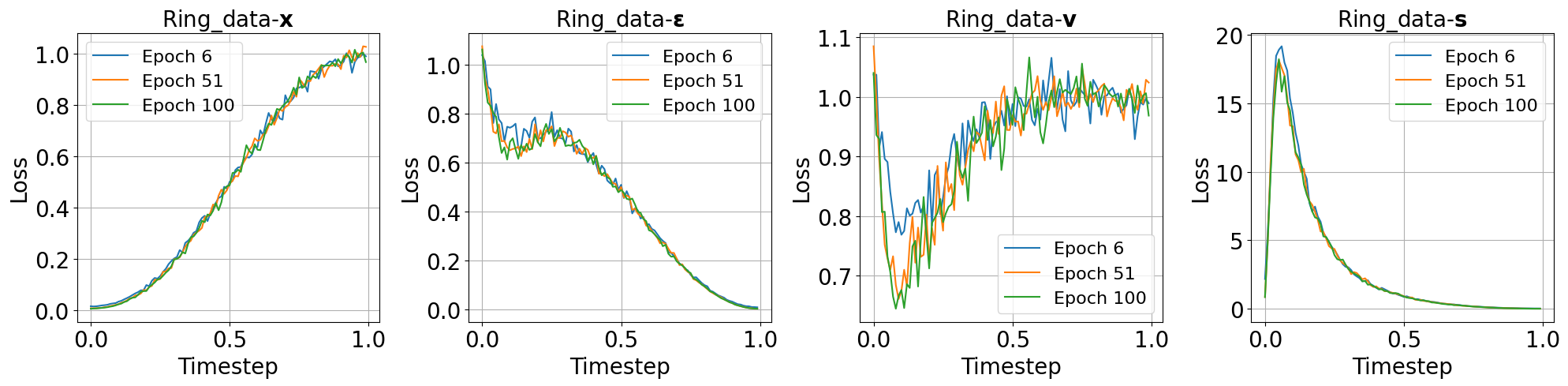}
    \caption{Behavior of weighted train loss with respect to timesteps for ring data at different epochs}
    \label{WTloss_vs_time_Ring}
\end{figure}

One of the reasons for this difference is the loss behavior of various target predictions across timesteps. We visualize the weighted train loss across timesteps for all target predictions on the ring dataset, as shown in Figure (\ref{WTloss_vs_time_Ring}). Similar trends are observed for other datasets, with corresponding graphs provided in the appendix C.3. In the $\rvx$-space, the model predicts the original data point at each timestep during the reverse diffusion. As noise increases in the forward process fig. \ref{2D_data_noise}, the $\text{SNR}$ drops significantly, making prediction harder and resulting in higher losses at later timesteps. In contrast, for $\rvepsilon$-space, the task is to predict the noise that was added to the data at each timestep. As more noise is introduced, predicting it becomes progressively easier. The $\rvv$-space formulation as shown in section \ref{vspace} interpolates between data $\rvx$ and noise $\rvepsilon$, weighted by time dependent functions, requiring the model to find a balance between the two. In $\rvs$-space the loss is significantly higher in the starting timesteps due to the sensitivity of score matching to noise variance ($\sigma_t$), as seen in equation \eqref{score_gaussian}. At early timesteps, $\sigma_t^2$ becomes negligible and the score function becomes very large in magnitude as $\rvs \propto \frac{1}{\sigma^2}$, leading to a significant rise in the loss.

\subsection{Results on image dataset}
The results of different loss formulations in the image dataset are presented in table \ref{tab:images_result}. 
We do not include results for score-based metrics because accurately computing them for continuous-time diffusion models in high-dimensional image space requires modeling reverse and forward Stochastic Differential Equations (SDEs), which is beyond the scope of this study and left for future research.

To evaluate the models, we used the NELBO loss to measure how well the model approximates the data likelihood, and the Frechet Inception Distance (FID) to assess the quality of generated samples, which were produced using 500 reverse diffusion steps.
We found that the NELBO formulation in $\rvx$-space has the best performance both in sample quality and probability density estimation. For $\rvepsilon$ and $\rvv$ space we found that the weighted loss formulation has better FID scores compared to NELBO. This again indicates that a more accurate likelihood estimation does not necessarily correspond to better sample quality.
The images generated from these experiments are provided in the appendix D.

\begin{table}[h!]
\centering
\caption{Comparison of NELBO and weighted loss formulations for CIFAR10}
\begin{tabular}{|p{1.8cm}|p{0.8cm}|p{1.1cm}|p{1cm}|p{1cm}|p{1cm}|}
\hline
\centering{\textbf{Data}} & \centering{\textbf{Loss Form}} & \multicolumn{2}{c|}{\textbf{\makecell{NELBO Loss \\ ($L$)}}} & \multicolumn{2}{c|}{\textbf{\makecell{Weighted loss \\ ($\mathcal{L}$)}}} \\ \cline{3-6} 
& & \textbf{Nelbo$\downarrow$} & \textbf{FID$\downarrow$} & \textbf{Loss$\downarrow$} & \textbf{FID$\downarrow$} \\ \hline
\centering{CIFAR10} & \centering{$\rvx$} & 0.0907 & 17.35 & 0.0544 & 19.08\\ 
& \centering{$\rvepsilon$} & 0.8499 & 39.77 & 0.0605 & 20.03\\ 
& \centering{$\rvv$} & 0.8576 & 36.84 & 0.1188 & 31.21\\ \hline
\end{tabular}
\label{tab:images_result}
\end{table}

\section{Conclusion}
In this work, we explored both the theoretical foundations and empirical behavior of various target prediction in diffusion models, with a focus on their corresponding loss formulations under the NELBO and weighted loss frameworks. By systematically deriving and relating the loss functions for different target predictions, that is data $\rvx$, noise $\rvepsilon$, rate of change of data distribution $\rvv$, and score $\rvs$, we established a unified understanding of how these objectives are connected at a theoretical level. 

We designed experiments to evaluate whether the mathematical equivalence of these objectives translates into similar empirical performance. Our results show that, despite theoretical equivalence, practical performance can differ significantly in certain scenarios. In particular, we observed variation in loss convergence, likelihood estimation, and sample quality across loss formulations. Among the NELBO variants, the formulation in the $\rvx$-space yielded the best likelihood estimates. The quality of generated samples was found to be comparable across both NELBO and weighted loss formulations in most cases for 2D datasets.
In contrast, for image data, the weighted loss showed improved performance in the $\rvepsilon$ and $\rvv$-spaces.

While our analysis is primarily conducted on 2D synthetic datasets, the insights gained offer a foundation for more extensive experiments on high dimensional image data. These findings highlight the importance of the choice of training objective in diffusion models and its impact on both model performance and sample quality. Overall, our study provides insights into the practical consequences of loss formulations and lays the groundwork for further research on optimizing training objectives in diffusion models.

\begin{credits}
\subsubsection{\ackname} This research is supported by the Center for Artificial Intelligence (CAIRO) at the Technical University of Applied Sciences Würzburg-Schweinfurt (THWS), Würzburg, Germany.

\subsubsection{\discintname}
The authors have no competing interests to declare that are relevant to the content of this article.
\end{credits}
%


\section*{Appendix}
\appendix

\section{Distributions}  
\label{sec:appendix_distributions}

The distribution for the forward and reverse diffusion process are defined by,

\begin{equation}
\begin{aligned}
q(\rvz_t \mid \rvx) &= \mathcal{N}(\rvz_t; \alpha_t \rvx, \sigma_t^2 \mI)
\end{aligned}
\label{q_z_x}
\end{equation}
\begin{equation}
\begin{aligned}
q(\rvz_t \mid \rvz_s) &= \mathcal{N}(\rvz_t; \alpha_{t|s} \rvz_s, \sigma_{t|s}^2 \mI)
\end{aligned}
\label{q_z_z}
\end{equation}

where $0 \leq s \leq t \leq 1$, $\alpha_{t \mid s} = \frac{\alpha_t}{\alpha_s}$ and $\sigma_{t \mid s}^2 = \sigma_t^2 - \alpha_{t \mid s}^2 \sigma_s^2$

\begin{equation}
    q(\rvz_s|\rvz_t, \rvx) = \mathcal{N} \left( \rvmu_Q(\rvz_t, \rvx; s, t), \sigma^2_Q(s, t)\mI \right)
\end{equation}
where
\begin{equation}
    \sigma^2_Q(s, t) = \frac{\sigma^2_{t|s} \sigma^2_s}{\sigma^2_t}
\end{equation}
and
\begin{equation}
    \rvmu_Q(\rvz_t, \rvx; s, t) = \alpha_{t|s} \frac{\sigma_s^2}{\sigma_t^2} \rvz_t + \alpha_s \frac{\sigma^2_{t|s}}{\sigma_t^2} \rvx
    \label{eq:mu_Q}
\end{equation}

The conditional model distribution is then chosen to be,

\begin{equation}
p_{\vtheta}(\rvz_s|\rvz_t) = q(\rvz_s \mid \rvz_t, \rvx = \hat{\rvx}_{\vtheta}(\rvz_t; t))
\end{equation}

where $\hat{\rvx}_\vtheta(\rvz_t; t)$ is the prediction of the original data $\rvx$ by our denoising model given the noisy data $\rvz_t$ and time t.

This implies,
\begin{equation}
p_{\vtheta}(\rvz_s|\rvz_t) = \mathcal{N} \left( \rvmu_{\vtheta}(\rvz_t; s, t), \sigma^2_Q(s, t)\mI \right)
\end{equation}

\section{Loss formulations}

\subsection{$\rvx$-space}
\label{apdx:xspace}

The diffusion loss is given as 

\begin{equation}
L_T(\rvx) = \sum_{i=1}^T \mathbb{E}_{q(\rvz_{t(i)} \mid \rvx)} \
D_{\text{KL}}\left[q(\rvz_{s(i)} \mid \rvz_{t(i)}, \rvx) \, \| \, p_\vtheta(\rvz_{s(i)} \mid \rvz_{t(i)})\right]
\label{eq:diff_loss}
\end{equation}

where for finite steps T, $s(i) = \frac{i - 1}{T}$ and $t(i) = \frac{i}{T}$. For simplicity, we will use short notations for $s(i)$ and $t(i)$. 

As the distributions $q(\rvz_s \mid \rvz_t, \rvx)$ and $p_\vtheta(\rvz_s \mid \rvz_t)$ are Gaussian with same variances, the KL divergence between these distributions has a closed form solution that simplifies to,

\begin{equation}
D_{\text{KL}}(q(\rvz_s \mid \rvz_t, \rvx) \parallel p(\rvz_s \mid \rvz_t)) = \frac{1}{2\sigma^2_Q(s, t)} \Vert \rvmu_Q - \rvmu_{\vtheta} \Vert^2
\end{equation}

Using equation \eqref{eq:mu_Q} we get,

\[ D_{KL}= \frac{\sigma_t^2}{2\sigma^2_{t|s}\sigma_s^2} \frac{\alpha_s^2\sigma_{t|s}^4}{\sigma_t^4} \Vert \rvx - \hat{\rvx}_{\vtheta}(\rvz_t; t) \Vert^2 \]

\[ = \frac{1}{2\sigma_s^2} \frac{\alpha_s^2\sigma_{t|s}^2}{\sigma_t^2} ||\rvx - \hat{\rvx}_\vtheta(\rvz_t; t)||_2^2 \]

\[ = \frac{1}{2\sigma_s^2} \frac{\alpha_s^2(\sigma_t^2 - \alpha_{t|s}^2\sigma_s^2)}{\sigma_t^2} ||\rvx - \hat{\rvx}_\vtheta(\rvz_t; t)||_2^2 \]

\[= \frac{1}{2} \frac{\alpha_{s}^{2} \sigma_{t}^{2} / \sigma_{s}^{2} - \alpha_{t}^{2}}{\sigma_{t}^{2}} \, \| \rvx - \hat{\rvx}_{\vtheta}(\rvz_{t}; t) \|_{2}^{2} \]

\[ = \frac{1}{2}\left( \frac{\alpha_s^2}{\sigma_s^2} - \frac{\alpha_t^2}{\sigma_t^2} \right) ||\rvx - \hat{\rvx}_\vtheta(\rvx_t; t)||_2^2 \]

\begin{equation}
= \frac{1}{2} (\text{SNR}(s) - \text{SNR}(t)) ||\rvx - \hat{\rvx}_\vtheta(\rvx_t; t)||_2^2 
\label{eq:in_SNR}
\end{equation}

Substituting this in equation \eqref{eq:diff_loss} we get,

\begin{equation}
L_T(\rvx) = \frac{1}{2} \mathbb{E}_{\epsilon \sim \mathcal{N}(0, \mI)} \left[ \sum_{i=1}^{T} (\text{SNR}(s) - \text{SNR}(t)) \, \| \rvx - \hat{\rvx}_{\vtheta}(\rvz_{t}; t) \|_{2}^{2} \right]
\end{equation}

\begin{equation}
L_T(\rvx) = \frac{T}{2} \, \mathbb{E}_{\rvepsilon \sim \mathcal{N}(\vzero, \mI), i \sim \text{U}\{1, T\}} 
\big[(\text{SNR}(s) - \text{SNR}(t)) \, \| \rvx - \hat{\rvx}_\vtheta(\rvz_t; t) \|_2^2 \big]
\label{eq:finite_elbo_x}
\end{equation}

For continuous time model $T \to \infty$, we can express it as $T = \frac{1}{\tau}$. In that case $s = \tau(i-1)$ and $t = \tau i$. Therefore,

\begin{equation}
L(\rvx) = \frac{1}{2} \mathbb{E}_{\epsilon \sim \mathcal{N}(0, \mI), i \sim U\{1, T\}} \left[ \frac{\text{SNR}(t - \tau) - \text{SNR}(t)}{\tau} \, \| \rvx - \hat{\rvx}_{\vtheta}(\rvz_t; t) \|_{2}^{2} \right]
\label{eq:intermediate_infinite_t}
\end{equation}

As $T \to \infty$, $\tau \to 0$. Additionally, we can remove the
factor $\frac{1}{2}$ as it is a constant scaling factor that does not affect the optimization process in terms of finding the optimal parameters $\vtheta$, which simplifies the equation \eqref{eq:intermediate_infinite_t} to,

\begin{equation}
L(\rvx) = -\mathbb{E}_{\epsilon \sim \mathcal{N}(0, \mI), t \sim \mathcal{U}[0, 1]} \left[ \text{SNR}'(t) \, \| \rvx - \hat{\rvx}_{\vtheta}(\rvz; t) \|_2^2 \right]
\end{equation}

\begin{equation}
L(\rvx) = - \mathbb{E}_{\epsilon \sim \mathcal{N}(0, \mI)} \int_0^1 \text{SNR}'(t) \, \| \rvx - \hat{\rvx}_{\vtheta}(\rvz_t; t) \|_2^2 \, dt
\label{eq:Final_infinite_t}
\end{equation}

Here, $\hat{\rvx}_\vtheta(\rvz_t; t)$ is the prediction of denoising model given the noisy data $\rvz_t = \alpha_t\rvx + \sigma_t\rvepsilon$ and time $t$.

For cosine schedule $\alpha_t = \cos(\pi/2)t$ and $\sigma_t = \sin(\pi/2)t$, therefore the $\text{SNR}(t)$ is given as,

\begin{equation}
\text{SNR}(t) = \frac{\alpha_t^2}{\sigma_t^2}
\end{equation}

\begin{equation}
\text{SNR}(t) = \frac{\cos^2(\pi/2)t}{\sin^2(\pi/2)t}
\end{equation}

\begin{equation}
\text{SNR}'(t) = \frac{-\pi\cos(\pi/2)t}{\sin^3(\pi/2)t} = \frac{-\pi\alpha_t}{\sigma_t^3}
\end{equation}

\subsection{$\rvepsilon$-space}
\label{apdx:epsilonspace}

To derive the loss formulation in the $\rvepsilon$-space we use equation \eqref{eq:Final_infinite_t} and write $\rvx$ in terms of $\rvepsilon$ using $\rvz_t = \alpha_t\rvx + \sigma_t\rvepsilon$,

\begin{equation}
L(\rvepsilon) = - \mathbb{E}_{\epsilon \sim \mathcal{N}(0, \mI)} \int_0^1 \text{SNR}'(t) \, \| (\rvz_t - \sigma_t\rvepsilon)/\alpha_t - (\rvz_t-\sigma_t\hat{\rvepsilon}_{\vtheta}(\rvz_t; t))/\alpha_t \|_2^2 \, dt
\end{equation}

\begin{equation}
L(\rvepsilon) = - \mathbb{E}_{\epsilon \sim \mathcal{N}(0, \mI)} \int_0^1 \text{SNR}'(t) \, \| (\rvz_t - \sigma_t\rvepsilon)/\alpha_t - (\rvz_t-\sigma_t\hat{\rvepsilon}_{\vtheta}(\rvz_t; t))/\alpha_t \|_2^2 \, dt
\end{equation}

\begin{equation}
L(\rvepsilon) = - \mathbb{E}_{\epsilon \sim \mathcal{N}(0, \mI)} \int_0^1 \text{SNR}'(t) \frac{\sigma_t^2}{\alpha_t^2} \, \| \rvepsilon - \hat{\rvepsilon}_{\vtheta}(\rvz_t; t) \|_2^2 \, dt
\end{equation}

\begin{equation}
L(\rvepsilon) = -\mathbb{E}_{\epsilon \sim \mathcal{N}(0, \mI)} \int_0^1 \frac{\text{SNR}'(t)}{\text{SNR}(t)} \, \| \rvepsilon - \hat{\rvepsilon}_{\vtheta}(\rvz_t; t) \|_2^2 \, dt
\end{equation}

\begin{equation}
L(\rvepsilon) = - \mathbb{E}_{\epsilon \sim \mathcal{N}(0, \mI), t \sim \mathcal{U}[0, 1]} \left[ \frac{\text{SNR}'(t)}{\text{SNR}(t)} \, \| \rvepsilon - \hat{\rvepsilon}_{\vtheta}(\rvz; t) \|_2^2 \right]
\end{equation}

\subsection{$\rvv$-space}
\label{apdx:vspace}

In v-space, $\rvz_{\phi_t} = \cos(\phi_t) \rvx + \sin(\phi_t) \rvepsilon$, where $\phi_t = \arctan\left( \frac{\sigma_t}{\alpha_t} \right)$ or $\tan(\phi_t) = \frac{\sigma_t}{\alpha_t}$, this implies,

\begin{equation}
\cos(\phi_t) = \frac{\alpha_t}{\sqrt{\alpha_t^2 + \sigma_t^2}}
\end{equation}

\begin{equation}
\sin(\phi_t) = \frac{\sigma_t}{\sqrt{\alpha_t^2 + \sigma_t^2}}
\label{eq:sin_phi}
\end{equation}

To get $\rvv$ we take the derivative of $\rvz_{\phi_t}$ with respect to $\phi_t$.

\begin{equation}
\rvv_{\phi_t} = \frac{d \rvz_{\phi_t}}{d \phi_t} = \frac{d \cos(\phi_t) \rvx}{d \phi_t} + \frac{d \sin(\phi_t) \rvepsilon}{d \phi_t} = \cos(\phi_t) \rvepsilon - \sin(\phi_t) \rvx
\end{equation}

Rearranging $ \rvepsilon, \rvx, \rvv $,
\begin{equation}
\sin(\phi_t) \rvx = \cos(\phi_t) \rvepsilon - \rvv_{\phi_t}
\end{equation}

\begin{equation}
\sin(\phi_t) \rvx = \frac{\cos(\phi_t) (\rvz_{\phi_t} - \cos(\phi_t) \rvx)}{\sin(\phi_t)} - \rvv_{\phi_t}
\end{equation}

\begin{equation}
\rvx (\sin^2(\phi_t) + \cos^2(\phi_t)) = \cos(\phi_t) \rvz_{\phi_t} - \sin(\phi_t) \rvv_{\phi_t}
\end{equation}

\begin{equation}
\rvx = \cos(\phi_t) \rvz_{\phi_t} - \sin(\phi_t) \rvv_{\phi_t}
\label{eq:v_relation_to_x}
\end{equation}

Using equation \eqref{eq:Final_infinite_t} and \eqref{eq:v_relation_to_x}, we get

\begin{align}
L(\rvv) = - \mathbb{E}_{\epsilon \sim \mathcal{N}(0, \mI)} \int_0^1 \text{SNR}'(t) \, 
\Big\| & (\cos(\phi_t)\rvz_{\phi_t} - \sin(\phi_t)\rvv) \notag \\
& - (\cos(\phi_t)\rvz_{\phi_t} - \sin(\phi_t)\hat{\rvv}_{\vtheta}(\rvz_t; t)) \Big\|_2^2 \, dt
\end{align}

\begin{equation}
L(\rvv) = - \mathbb{E}_{\epsilon \sim \mathcal{N}(0, \mI)} \int_0^1 \text{SNR}'(t)\sin^2\phi_t \, \| \rvv - \hat{\rvv}_{\vtheta}(\rvz_t; t) \|_2^2 \, dt
\end{equation}

Substituting $\sin^2\phi_t$ from equation $\eqref{eq:sin_phi}$,

\begin{equation}
L(\rvv) = - \mathbb{E}_{\epsilon \sim \mathcal{N}(0, \mI)} \int_0^1 \frac{\sigma_t^2}{\alpha_t^2 + \sigma_t^2} \text{SNR}'(t) \, \| \rvv - \hat{\rvv}_{\vtheta}(\rvz_t; t) \|_2^2 \, dt
\end{equation}

\begin{equation}
L(\rvv) = - \mathbb{E}_{\epsilon \sim \mathcal{N}(0, \mI), t \sim \mathcal{U}[0, 1]} \left[ \frac{\sigma_t^2}{\alpha_t^2 + \sigma_t^2} \, \text{SNR}'(t) \, \| \rvv - \hat{\rvv}_\vtheta(\rvz_t; t) \|_2^2 \right]
\end{equation}

\subsection{$\rvs$-space}
\label{apdx:sspace}

We use Tweedie's formula to show the relevance of score and ELBO formulation. It is used to improve the estimates of true mean of an underlying distribution, when we have noisy samples. For a Gaussian variable \( \rvz \sim \mathcal{N}(\rvz; \boldsymbol{\mu}_z, \boldsymbol{\Sigma}_z) \), Tweedie's formula states that,

\begin{equation}
\mathbb{E}[\boldsymbol{\mu}_z | \mathbf{z}] = \mathbf{z} + \boldsymbol{\Sigma}_z \nabla_{\mathbf{z}} \log p(\mathbf{z})
\end{equation}

From equation \eqref{q_z_x} we know that,

\[q(\rvz_t \mid \rvx) = \mathcal{N}(\rvz_t; \alpha_t \rvx, \sigma_t^2 \mI)\]

We will use the Tweedie's formula to get the true posterior mean of $q(\rvz_t \mid \rvx)$, 

\begin{equation}
\alpha_t \rvx = \rvz_t + \sigma_t^2 \nabla \log q(\rvz_t|\rvx)
\end{equation}

\begin{equation}
\rvx = \frac{\rvz_t + \sigma_t^2 \nabla \log q(\rvz_t|\rvx)}{\alpha_t}
\end{equation}

Substituting this in equation \eqref{eq:Final_infinite_t},

\begin{equation}
L(\rvs) = - \mathbb{E}_{\epsilon \sim \mathcal{N}(0, \mI)} \int_0^1 \text{SNR}'(t) \left\| \frac{\rvz_t}{\alpha_t} + \frac{\sigma_t^2}{\alpha_t} \nabla \log q(\rvz_t|\rvx) - \frac{\rvz_t}{\alpha_t} - \frac{\sigma_t^2}{\alpha_t} \hat{\rvs}_\vtheta(\rvz_t; t) \right\|_2^2 \, dt
\end{equation}


\begin{equation}
L(\rvs) = - \mathbb{E}_{\epsilon \sim \mathcal{N}(0, \mI)} \int_0^1 \text{SNR}'(t) \frac{\sigma_t^4}{\alpha_t^2} \left\| \nabla \log q(\rvz_t|\rvx) - \hat{\rvs}_\vtheta (\rvz_t; t) \right\|_2^2 \, dt
\end{equation}

\begin{equation}
L(\rvs) = - \mathbb{E}_{\epsilon \sim \mathcal{N}(0, \mI), t \sim \mathcal{U}[0, 1]} \left[ \text{SNR}'(t) \frac{\sigma_t^4}{\alpha_t^2} 
\left\| \nabla \log q(\rvz_t|\rvx) - \hat{\rvs}_\vtheta (\rvz_t; t) \right\|_2^2 \right]
\end{equation}

\section{2D datasets}
\label{apdx:2D_data}

Below we present the generated samples from different number of sampling steps for ring data, cluster data, swiss roll data and waves data in the NELBO and weighted loss formulations.

\subsection{Generated samples for NELBO loss}
\label{apdx:2D_NELBO}

\begin{figure}[h!]
    \centering
    \includegraphics[width=10cm, height = 5cm]{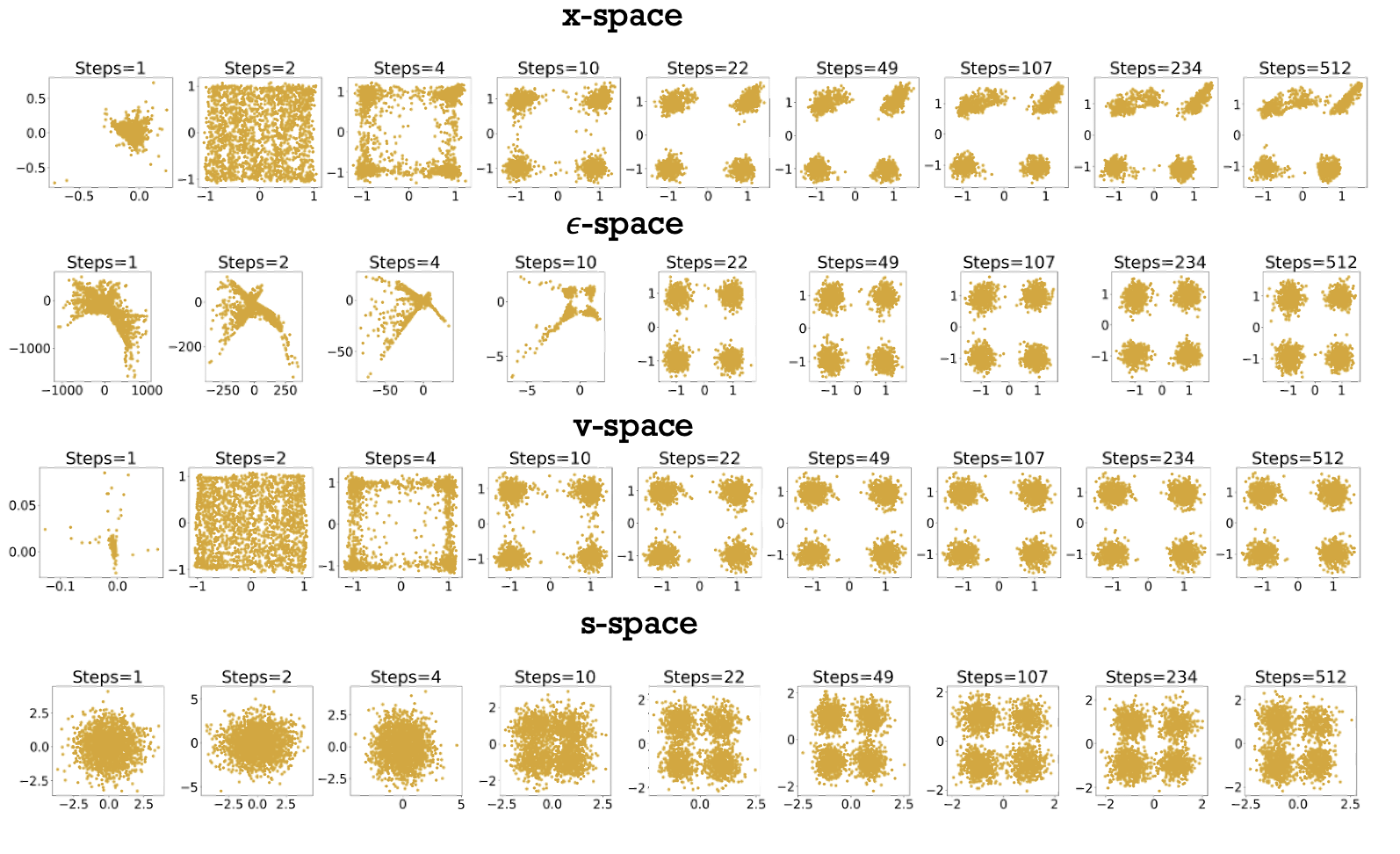}
    \caption{Generated samples for cluster data when using NELBO loss.}
\end{figure}

\begin{figure}[h!]
    \centering
    \includegraphics[width=10cm, height = 5cm]{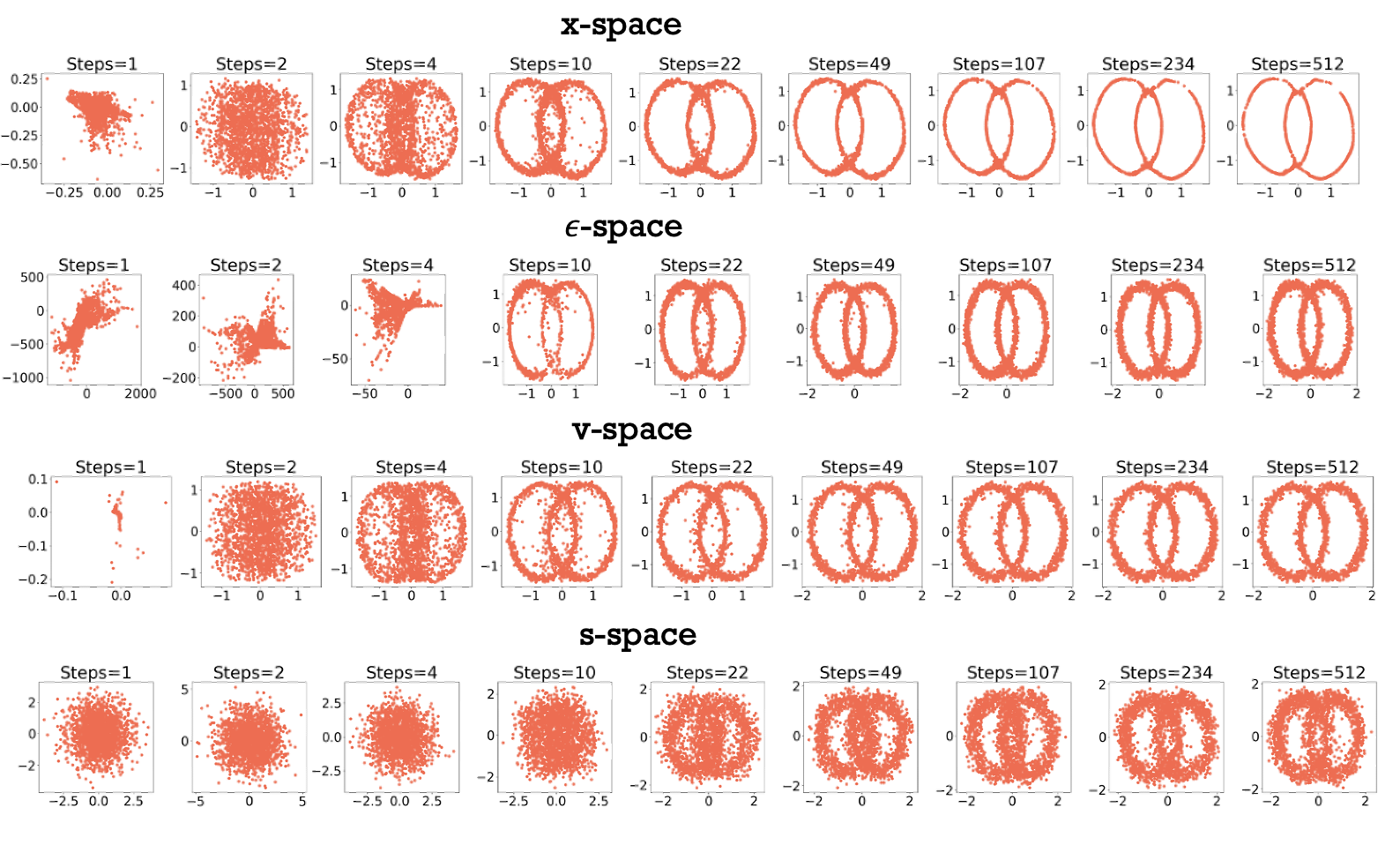}
    \caption{Generated samples for ring data when using NELBO loss.}
\end{figure}

\begin{figure}[h!]
    \centering
    \includegraphics[width=10cm, height = 5cm]{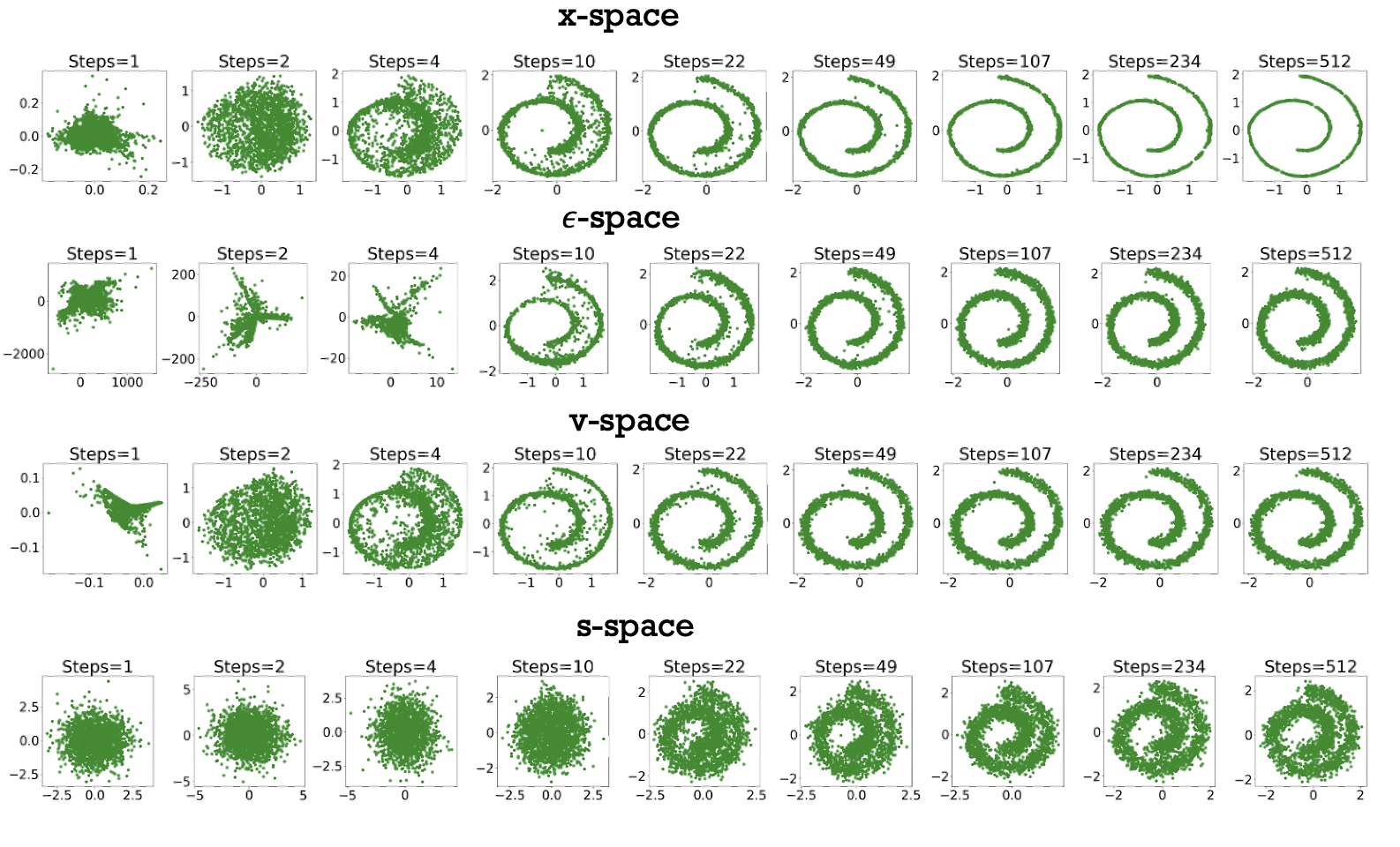}
    \caption{Generated samples for swiss roll data when using NELBO loss.}
\end{figure}

\begin{figure}[h!]
    \centering
    \includegraphics[width=10cm, height = 5cm]{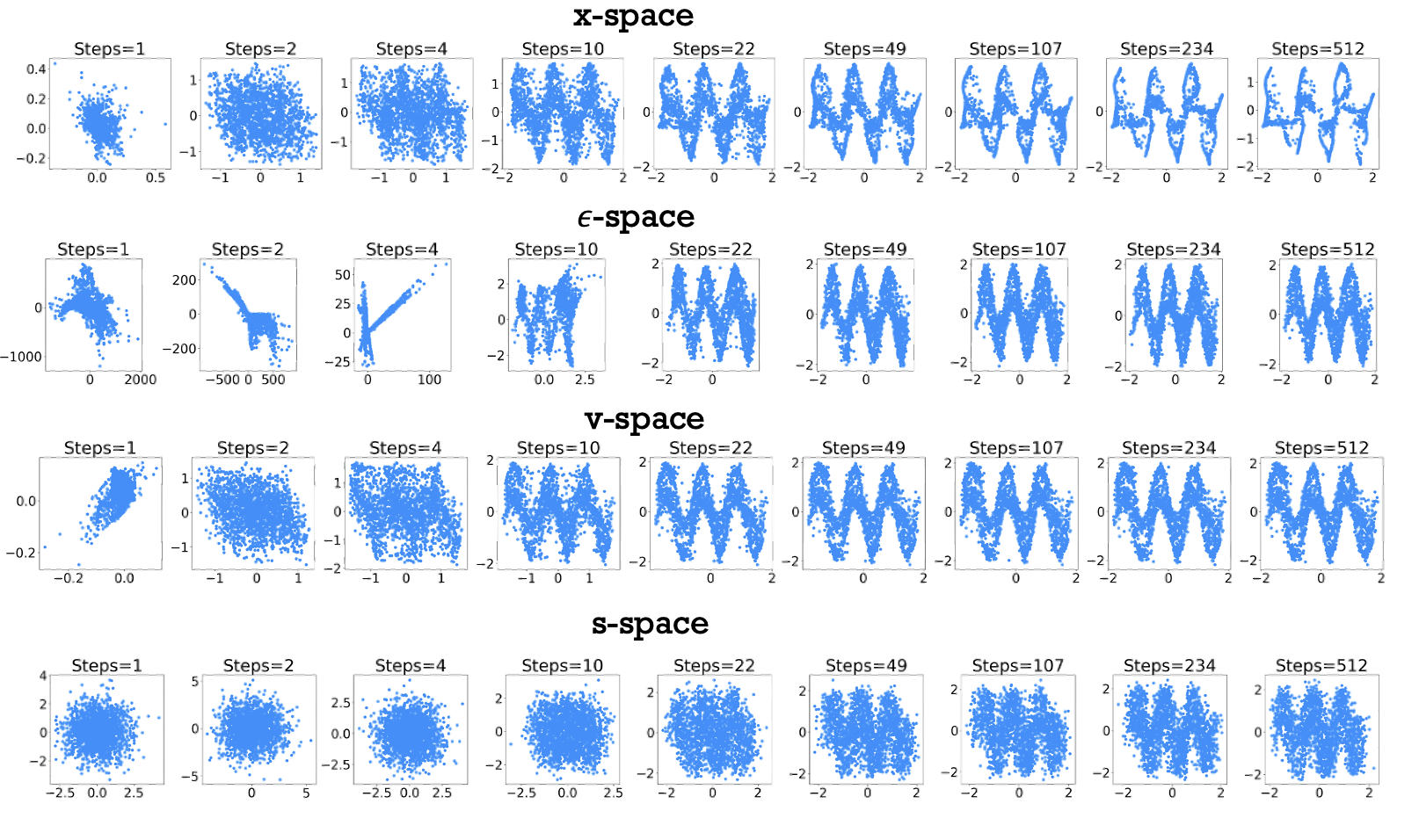}
    \caption{Generated samples for waves data when using NELBO loss.}
\end{figure}

\subsection{Generated samples for weighted loss}
\label{apdx:2D_WTD}

\begin{figure}[H]
    \centering
    \includegraphics[width=10cm, height = 5cm]{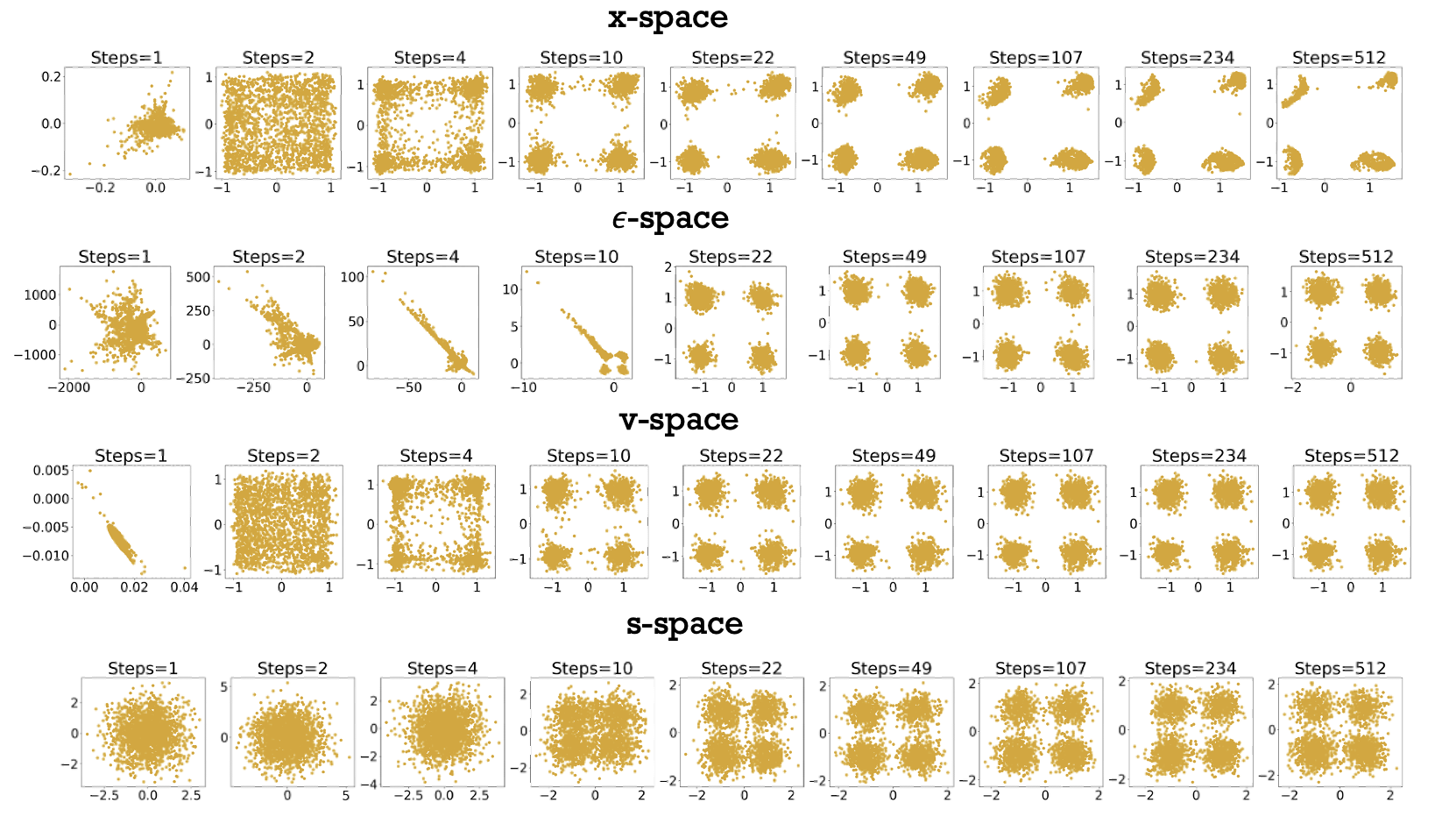}
    \caption{Generated samples for cluster data when using weighted loss.}
\end{figure}

\begin{figure}[H]
    \centering
    \includegraphics[width=10cm, height = 5cm]{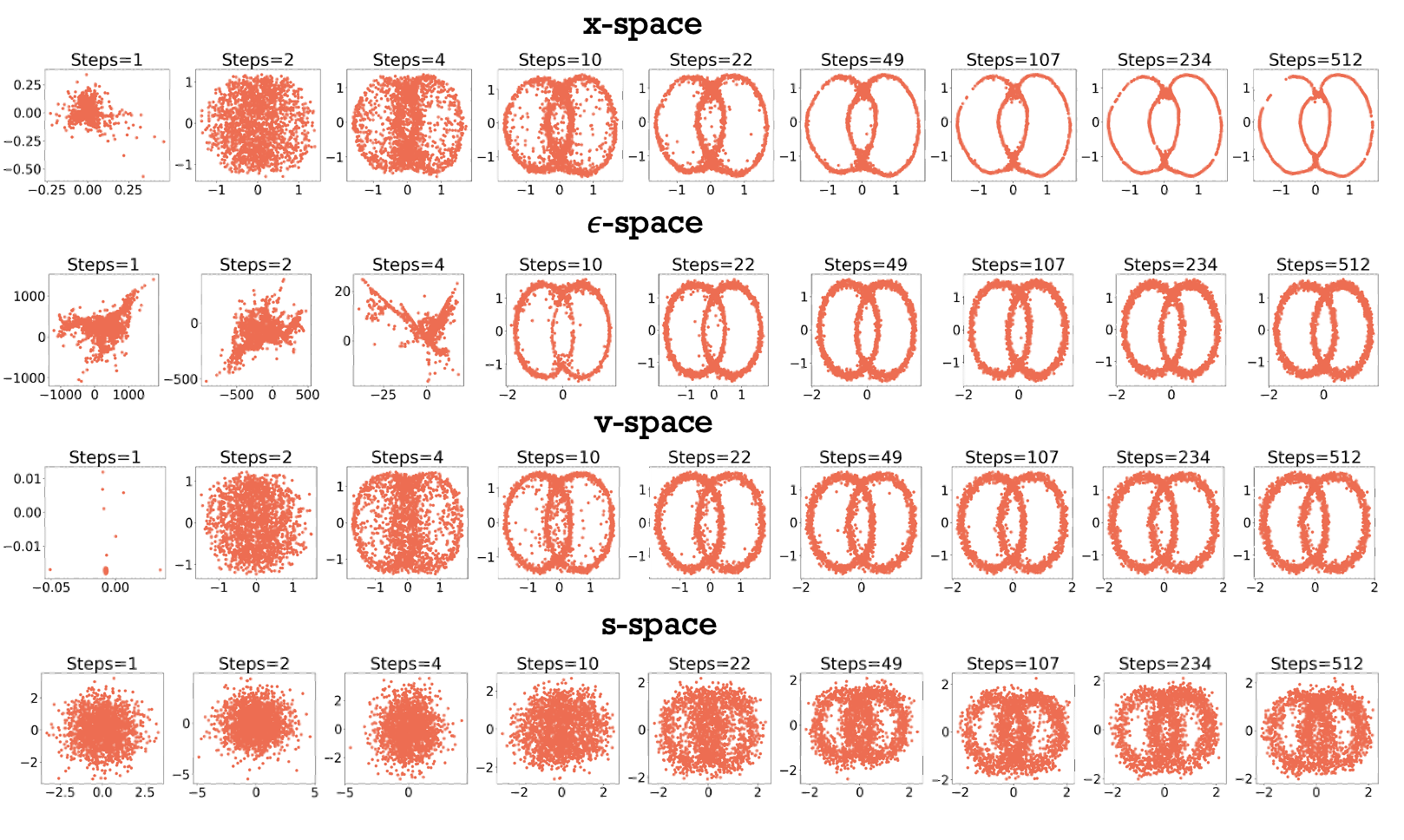}
    \caption{Generated samples for ring data when using weighted loss.}
\end{figure}

\begin{figure}[H]
    \centering
    \includegraphics[width=0.85\textwidth]{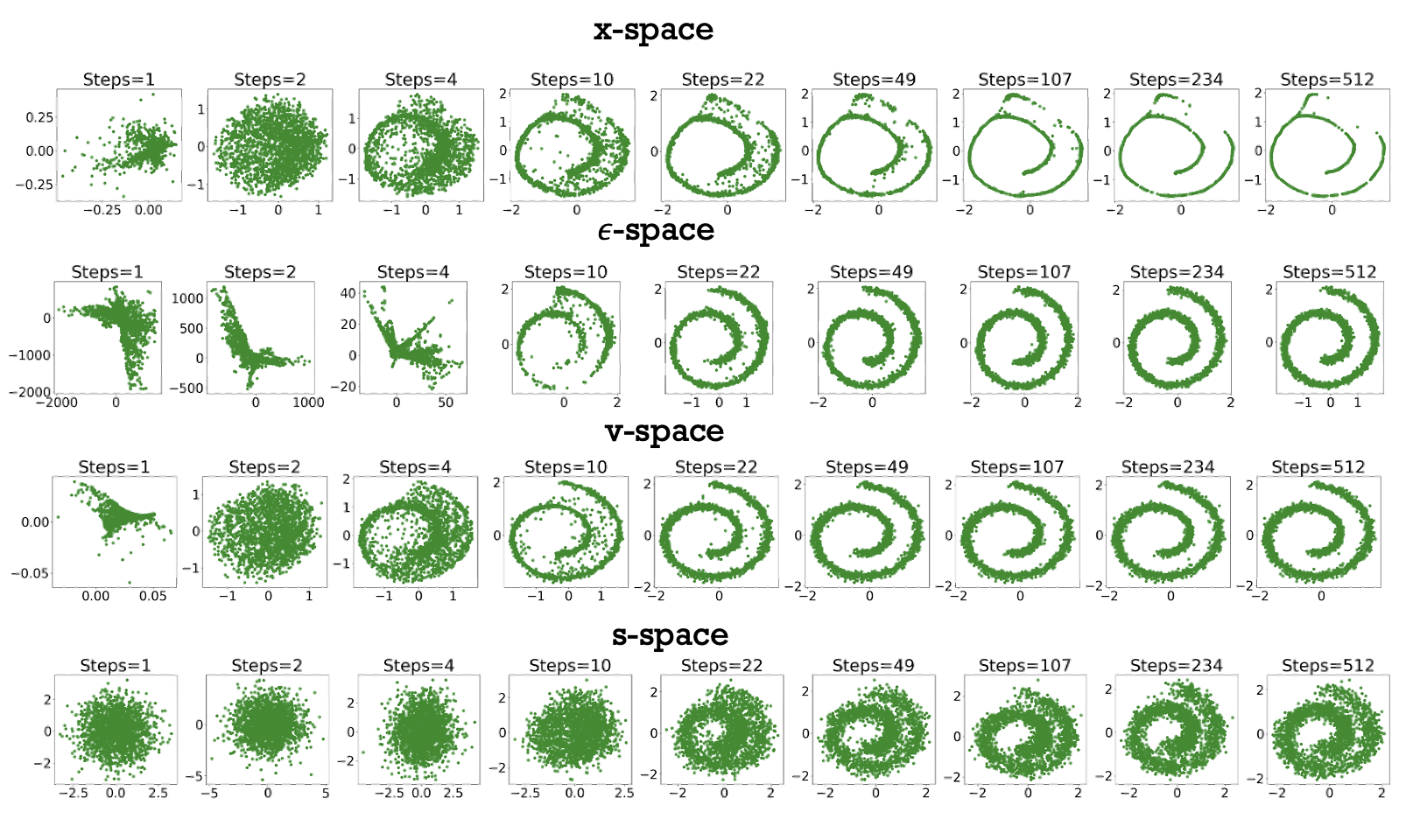}
    \caption{Generated samples for swiss roll data when using weighted loss.}
\end{figure}

\begin{figure}[H]
    \centering
    \includegraphics[width=0.85\textwidth]{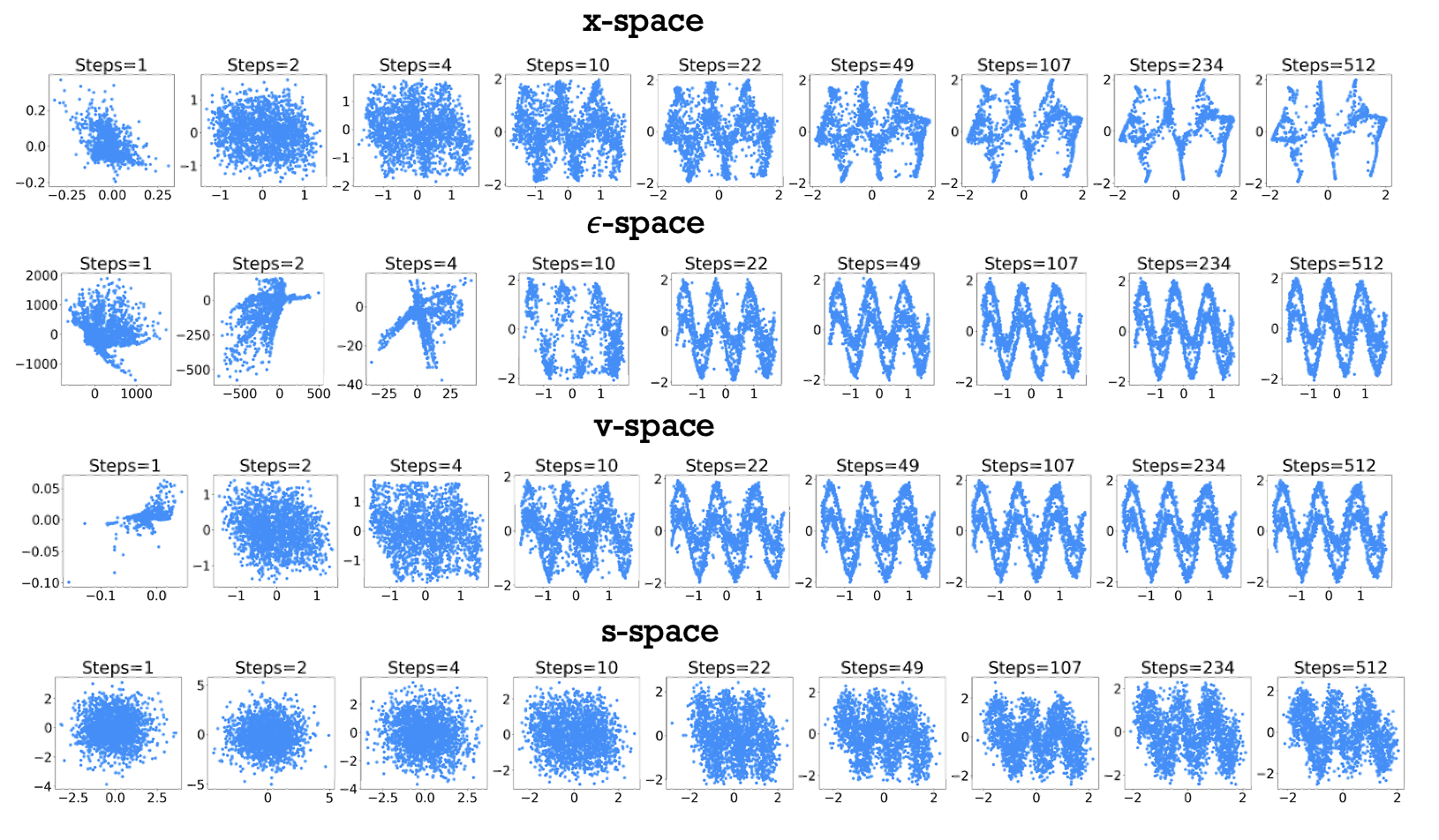}
    \caption{Generated samples for waves data when using weighted loss.}
\end{figure}

\newpage
\subsection{Weighted loss vs timestep}
\label{apdx:Loss_vs_T}

We show here how the weighted train loss behave with respect to time steps for other datasets.

\begin{figure}[H]
    \centering
    \includegraphics[width=12cm, height = 3.2cm]{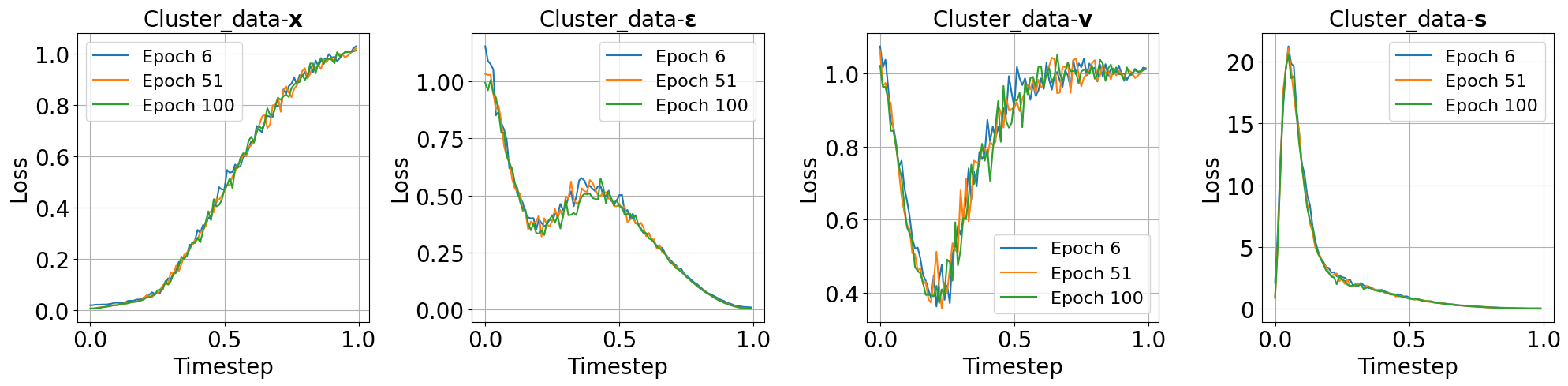}
    \caption{Weighted loss vs timestep for cluster data}
\end{figure}

\begin{figure}[H]
    \centering    
    \includegraphics[width=12cm, height = 3.2cm]{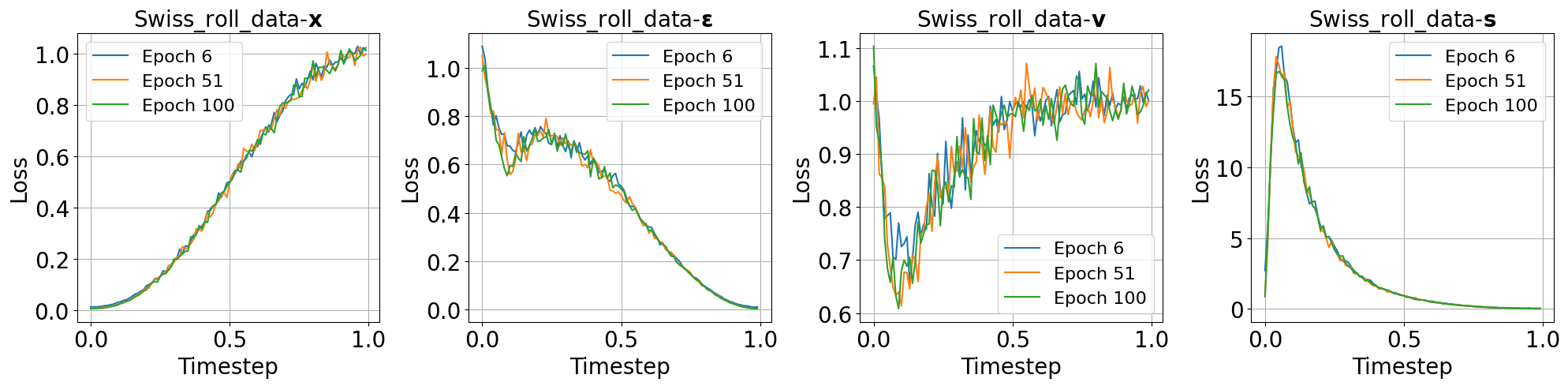}
    \caption{Weighted loss vs timestep for swiss roll data}
\end{figure}

\begin{figure}[H]
    \centering
    \includegraphics[width=12cm, height = 3.2cm]{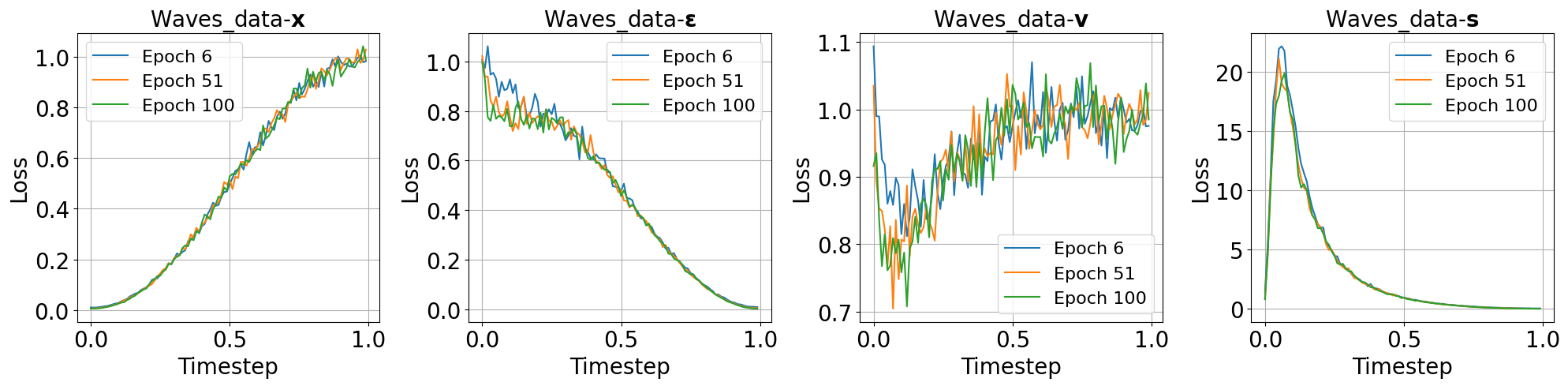}
    \caption{Weighted loss vs timestep for waves data}
\end{figure}

\section{Image dataset}
\label{apdx:Img_data}

\subsection{$\rvx$-space}

\begin{figure}[H]
    \centering
    \includegraphics[width=7cm, height = 7cm]{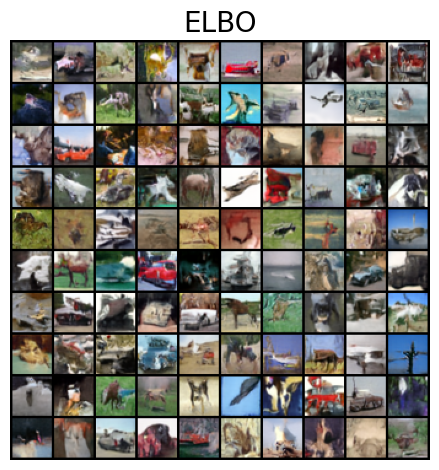}
    \caption{Images generated for NELBO $\rvx$ formulation}
\end{figure}

\begin{figure}[H]
    \centering
    \includegraphics[width=7cm, height = 7cm]{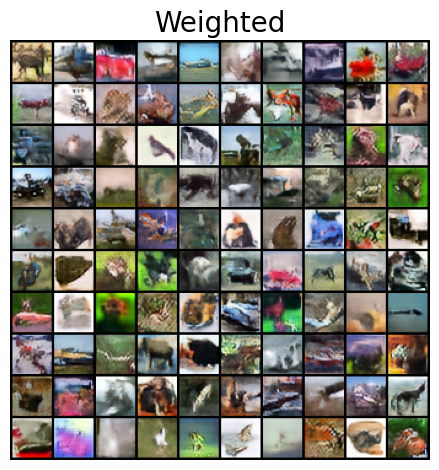}
    \caption{Images generated for weighted $\rvx$ formulation}
\end{figure}

\subsection{$\rvepsilon$-space}

\begin{figure}[H]
    \centering
    \includegraphics[width=7cm, height = 7cm]{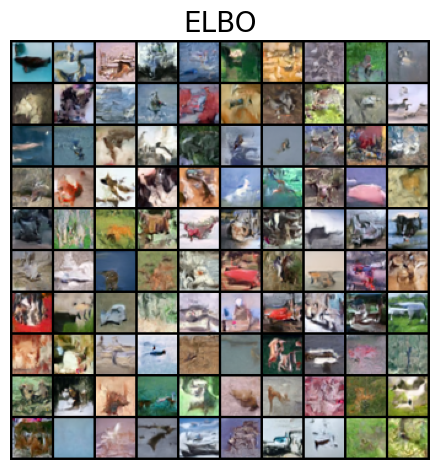}
    \caption{Images generated for NELBO $\rvepsilon$ formulation}
\end{figure}

\begin{figure}[H]
    \centering
    \includegraphics[width=7cm, height = 7cm]{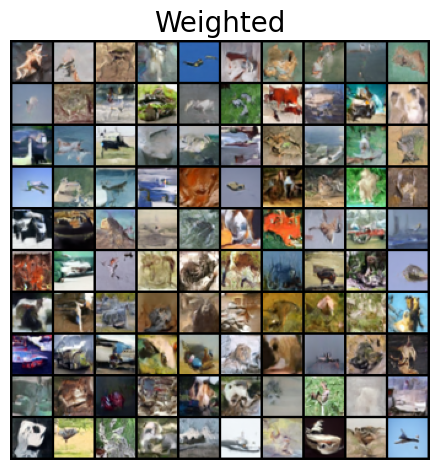}
    \caption{Images generated for weighted $\rvepsilon$ formulation}
\end{figure}

\subsection{$\rvv$-space}

\begin{figure}[H]
    \centering
    \includegraphics[width=7cm, height = 7cm]{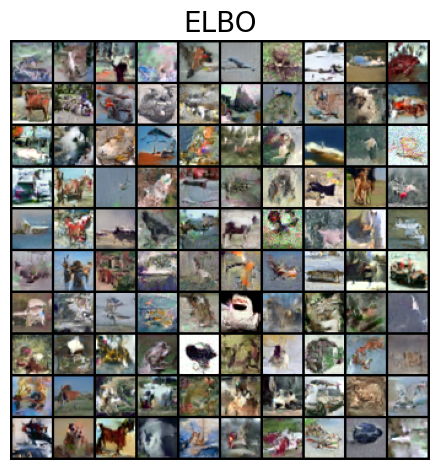}
    \caption{Images generated for NELBO $\rvv$ formulation}
\end{figure}

\begin{figure}[H]
    \centering
    \includegraphics[width=7cm, height = 7cm]{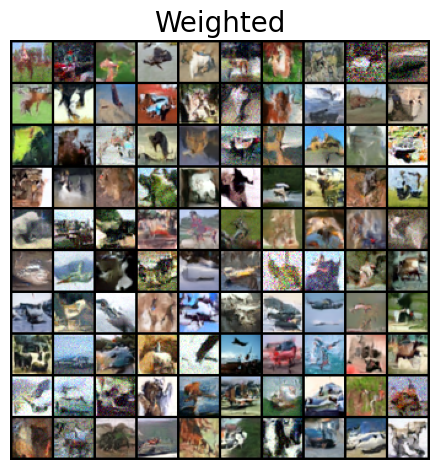}
    \caption{Images generated for weighted $\rvv$ formulation}
\end{figure}

\end{document}